\documentclass[lettersize,journal]{IEEEtran}
\usepackage{amsmath,amsfonts}
\usepackage{algorithmic}
\usepackage{amssymb}
\usepackage{algorithm}
\usepackage{array}
\usepackage[caption=false,font=normalsize,labelfont=sf,textfont=sf]{subfig}
\usepackage{textcomp}
\usepackage{stfloats}
\usepackage{url}
\usepackage{verbatim}
\usepackage{graphicx}
\usepackage{cite}

\begin{document}

\title{Modular Isoperimetric Soft Robotic Truss for Lunar Applications}

\author{Mihai Stanciu$^{1}$, Isaac Weaver$^{1}$, Adam Rose$^{1}$, James Wade$^{1}$, Kaden Paxton$^{1}$, Chris Paul$^{1}$, Spencer Stowell$^{1}$, Nathan Usevitch$^{1*}$, ~\IEEEmembership{Member,~IEEE}
\thanks{This work was supported by NASA Grant No. 80NSSC20M0103, the 2024 NASA BIG Idea Challenge, and NASA Grant No. 80NSSC26K0260}
\thanks{$^{1}$All authors are with the Ira A. Fulton School of Engineering, Mechanical Engineering Department, Brigham Young University, Provo, Utah, US}
\thanks{ $^{*}$ Corresponding Author {\tt\footnotesize nathan\_usevitch@byu.edu}, (801) 422-0814}}



\maketitle

\begin{abstract}
We introduce a large-scale robotic system designed as a lightweight, modular, and reconfigurable structure for lunar applications. The system consists of truss-like robotic triangles formed by continuous inflated fabric tubes routed through two robotic roller units and a connecting unit. A newly developed spherical joint enables up to three triangles to connect at a vertex, allowing construction of truss assemblies beyond a single octahedron. When deflated, the triangles compact to approximately the volume of the roller units, achieving a stowed-to-deployed volume ratio of 1:18.3. Upon inflation, the roller units pinch the tubes, locally reducing bending stiffness to form effective joints. Electric motors then translate the roller units along the tube, shifting the pinch point by lengthening one edge while shortening another at the same rate, thereby preserving a constant perimeter (isoperimetric). This shape-changing process requires no additional compressed air, enabling untethered operation after initial inflation. We demonstrate the system as a 12-degree-of-freedom solar array capable of tilting up to $60^{\circ}$ and sweeping 
$360^{\circ}$, and as a 14-degree-of-freedom locomotion device using a step-and-slide gait. This modular, shape-adaptive system addresses key challenges for sustainable lunar operations and future space missions.
\end{abstract}


\section{Introduction}
\IEEEPARstart{E}{stablishing} a permanent human presence on the moon represents a key goal for space exploration \cite{nasa2022strategicplan}. To ensure that lunar outposts can operate efficiently, there is a need for adaptable structures and robots that can interact safely with astronauts or complete tasks autonomously. To address this challenge, we have developed an isoperimetric inflatable robotic truss designed to function as a reconfigurable structure, capable of adapting its shape to perform various tasks suited for lunar missions. This robotic structure has 14 active degrees of freedom, in some configurations stands over 3.5 m high, and can be configured to various shapes. Such an inflatable soft robotic structure offers significant advantages over traditional structures, including its lightweight design and compact stowed volume, both critical factors for space travel. Its compliance and modularity make it well-suited for environments that require reliable and adaptable structures that can operate safely alongside humans. On the Moon, where safety, space efficiency, and versatility are critical, this modular isoperimetric inflatable robot provides a valuable solution.

\begin{figure}[htbp]
  \centering
  \includegraphics[width=.95\linewidth]{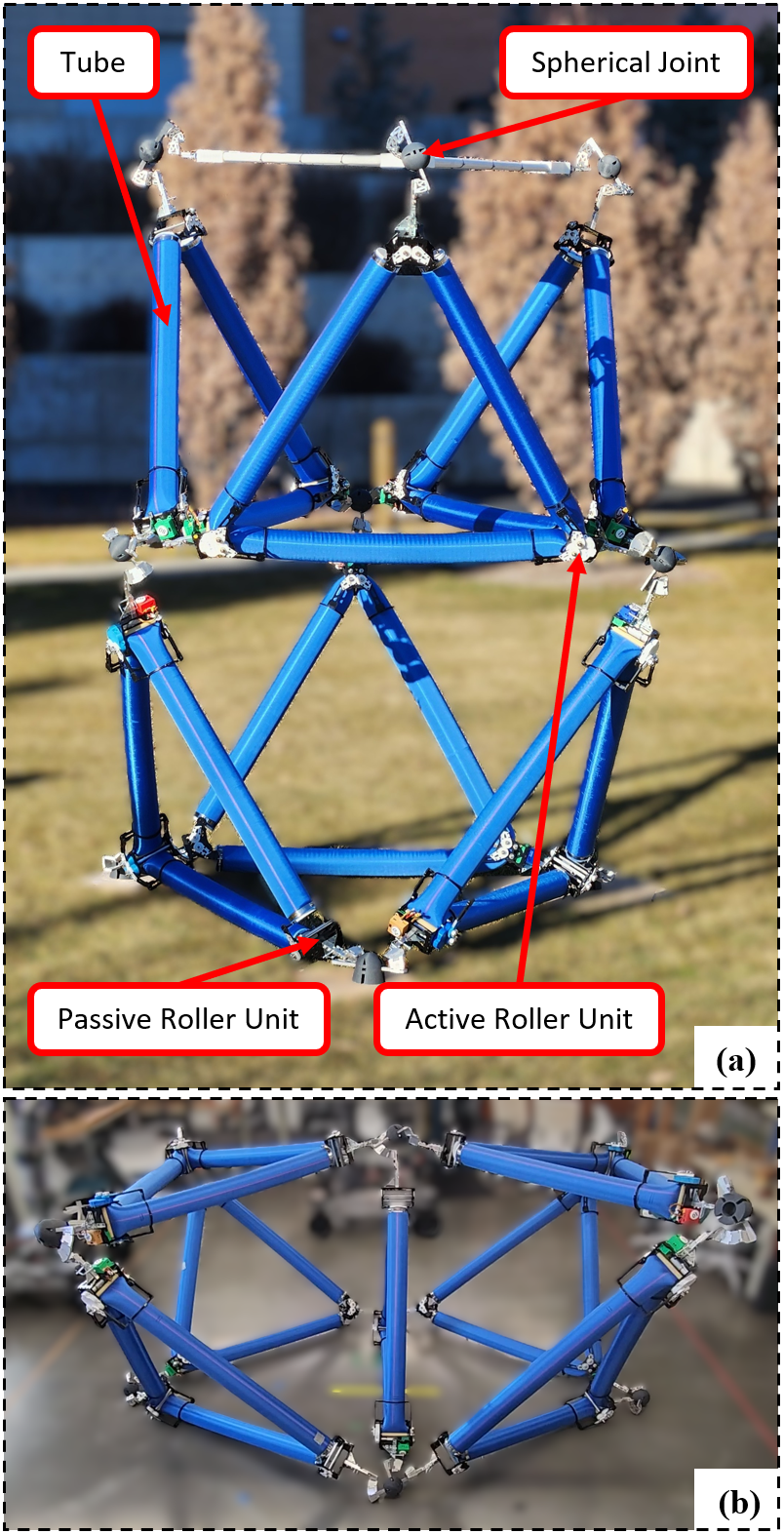}
  \caption{The isoperimetric truss robot with its main subsystems highlighted: the inflated tubes, the spherical joints, and the active and passive roller units. a) Solar array configuration. b) Locomotion configuration.}
  \label{octahedron_with_annotations}
\end{figure}

As shown in Fig.~\ref{octahedron_with_annotations}a, the robot consists of inflatable fabric tubes that are bent into triangles and are connected at spherical joints to form a truss structure. The active roller units, positioned at two vertices of each triangle, are capable of translating along the tubes, while the passive unit connects the two ends of the tube to complete the triangle. As the active units traverse the tubes, they change the side lengths of each triangle, thereby reshaping the overall configuration of the truss while preserving its perimeter. A specialized spherical joint, designed for this robot, connects the roller units of adjacent triangles at each vertex, allowing the construction of the truss. This paper discusses two configurations of the truss robot, specifically the solar array illustrated in Fig.~\ref{octahedron_with_annotations}a and the locomotion configuration illustrated in Fig.~\ref{octahedron_with_annotations}b. The solar array is composed of 6 triangles (12 controllable degrees of freedom) forming two vertically stacked octahedra that share a middle virtual triangle and an equilateral static solar panel at the top. The locomotion configuration is composed of 7 triangles (14 controllable degrees of freedom) that form two symmetric octahedra about a common middle triangle.

\subsection{Related Works}

\subsubsection{Lunar Structures}

Significant research has been conducted to develop lunar structures that would enable astronauts to perform tasks during extended expeditions on the Moon. Several studies have since been published on lunar structures, including habitats \cite{1_benaroya_lunar_habitats,2_ruess_lunar_structures}, solar arrays \cite{3_halbach_solar_array}, and formations designed to support scientific data collection \cite{4_bandyopadhyay_lunar_telescope}. These structures include monolithic continuous designs constructed using lunar materials such as regolith \cite{5_cesaretti_3Dprinted_regolith}, multi-member structures such as trusses \cite{6_nunan_lunar_truss}, deployable structures \cite{7_brennan_deployable_habitat}, and inflatable structures \cite{gruber2007deployable, wang2024novel}.

Truss, deployable, and inflatable structures are preferred for lunar applications and space travel because of their lightweight design, compact storage, and modularity. Due to these benefits, lunar astronauts tasked with constructing such truss structures would spend minimal time assembling them \cite{2_ruess_lunar_structures}.

\subsubsection{Lunar Robots}

Robotic systems for lunar applications have been under development for several decades. These systems offer promising alternatives to human astronauts for construction and maintenance tasks on lunar bases \cite{8_austin_RLSO2}, while also enabling new opportunities in exploration \cite{9_miaja_robocrane,10_vaquero_EELS}, sample collection and delivery \cite{11_eich_multirobot_missions}, and terrain mapping \cite{12_park_lunar_mapping}. Across all applications, lunar robots are designed to reduce risks and resource requirements for human personnel while expanding the operational capabilities of lunar missions.

A primary challenge in implementing lunar robots lies in their high degree of specialization. The literature contains numerous designs tailored to specific tasks; however, most exhibit limited adaptability beyond their intended functions, highlighting the need for more versatile robotic systems. One example addressing this limitation is the multi-robot system presented in \cite{11_eich_multirobot_missions}, which adapts by employing multiple robot configurations to perform sampling operations, first surveying and planning the terrain and then collecting and handling samples. Another major challenge is ensuring the successful delivery and operation of these systems on the lunar surface. As reported in \cite{13_chien_exploring_using_space_robots}, nearly half of the lunar robotic missions launched since 2019 have failed due to crashes or loss of communication. These issues underscore the need for modular robotic systems that, once delivered to the Moon, can be reconfigured to perform a wide range of tasks required at a lunar base.

\subsubsection{Truss Robots}

Several concepts and designs for reconfigurable truss robots, commonly referred to as variable geometry trusses (VGTs), have been proposed in the literature \cite{rhodes1985deployable, miura_variable_1985, spinos_topological_2021}, with some specifically developed for space-related applications \cite{hamlin1996tetrobot, curtis2007tetrahedral}. These systems are typically composed of rigid linear actuators capable of extending and retracting to modify the truss geometry. While such designs provide high structural stiffness and controllable kinematics, their reliance on rigid members and linear actuators introduces actuator stroke limits, stowability constraints, and increased safety risks from high-impact collisions with humans. These limitations restrict VGTs’ achievable geometries, continuous deformation, and safety, inhibiting their adaptability, compact transport potential, and safe operation around humans.

Usevitch et al. \cite{14_usevitch_untethered} introduced an untethered soft robotic truss that replaces rigid members with isoperimetric pressurized tubes. By employing isoperimetric tubes, this architecture preserves the geometric controllability of traditional VGTs, enables a broader range of motions than is achievable with purely linear actuators, and leverages the compliance and lightweight nature of soft structures. Hancey et al. \cite{hancey2026isoperimetric} proposed an alternative implementation of isoperimetric members using tape-spring elements, which behave as stiff beams along their length but become locally compliant when pinched at the vertices. Both designs demonstrate the concept for shape reconfiguration through mechanical actuation operating on isoperimetric members, highlighting the potential for deployable and resilient truss architectures. However, \cite{14_usevitch_untethered, hancey2026isoperimetric} only demonstrate the octahedron, primarily focusing on the feasibility of the isoperimetric approach rather than extending its functionality or modularity. 

\subsection{Contributions}

Building on the foundation established in \cite{14_usevitch_untethered}, our work extends the isoperimetric truss concept to incorporate new functionalities and demonstrate novel applications in the context of lunar missions. The primary contributions of our work are as follows:

\begin{itemize}
    \item Redesigned the robot to be modular, easily assembled in a lunar environment, and capable of compact stowage with a high stowed-to-deployed volume ratio suitable for space transportation.

    \item Developed a novel spherical joint enabling the connection of three or more components at a node while maintaining a virtual center of rotation.

    \item Demonstrated scalable structures beyond a single octahedron by constructing a stacked octahedron and validating its ability to perform complex motions, including extension, twisting, top-plane tilting and sweeping, and locomotion.


    \item Improved performance and reliability over prior work \cite{14_usevitch_untethered} by using high-pressure tubing, strengthening and minimizing the weight of the roller units, and implementing a more robust gear train for the active units, substantially increasing load capacity and actuation reliability.
\end{itemize}


\section{Methods}

\subsection{Design}

We first describe the methods used to construct the robot’s various subsystems, including the roller units, tubes, and spherical joints, as well as the modeling techniques employed to characterize its kinematics and dynamics.

\subsubsection{Roller Units}

The roller units form the subsystem responsible for bending and shaping the tubes into the desired configurations. The system uses two types of roller units: active rollers, which are motorized to traverse the tube, and passive rollers, which connect the two ends of the tube to form a triangle.

\begin{figure}[tbp]
  \centering
  \includegraphics[width=\linewidth]{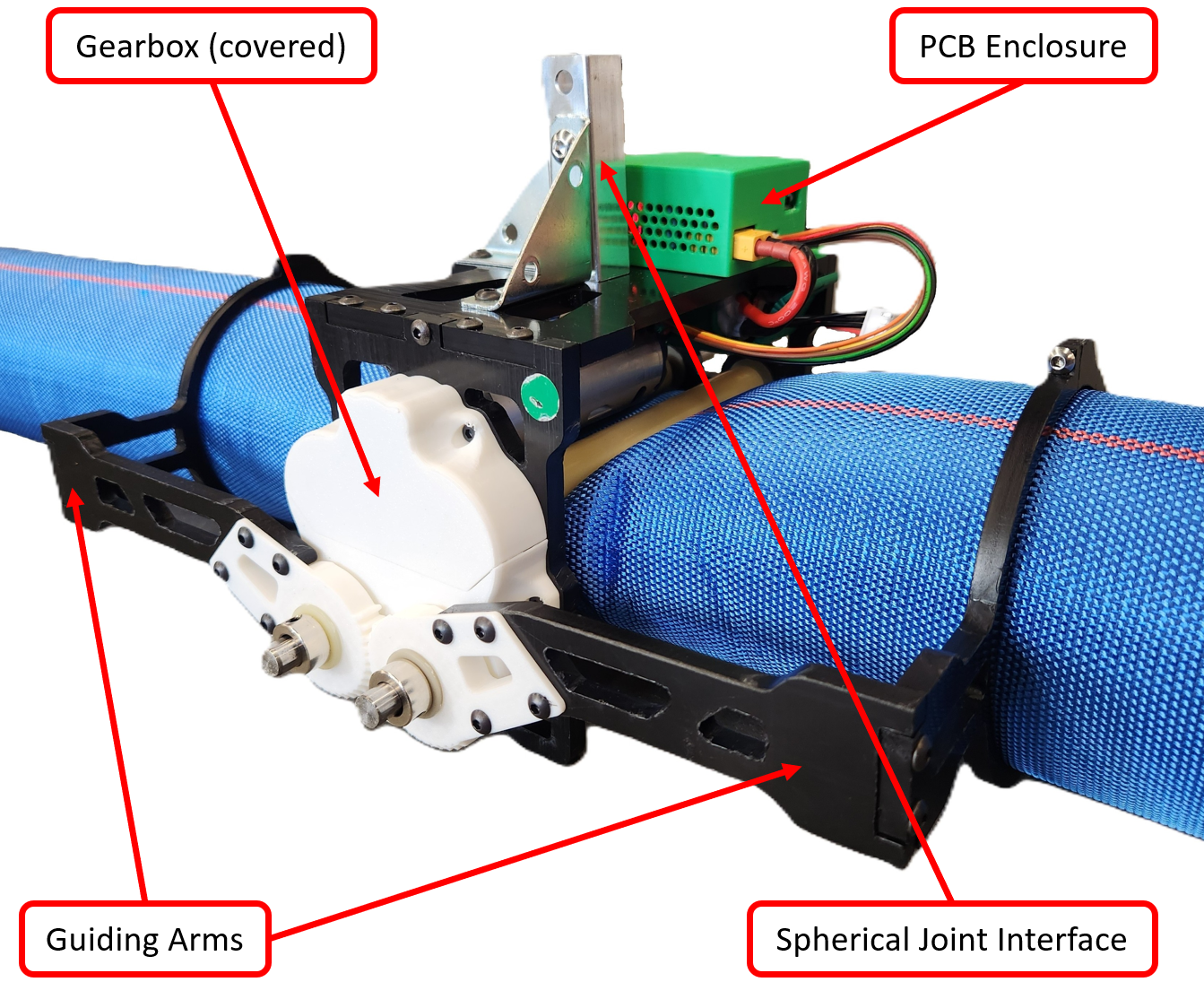}
  \caption{Active roller unit showcasing its external components.}
  \label{active_roller_unit}
\end{figure}

Both types of roller units must be lightweight to reduce load on the structure and strong enough to withstand loads through the spherical joint interface. The active unit must also sustain the tension forces created by the internal pressure of the tubes as it acts on the roller shafts. The passive roller unit must withstand stress from the spring-back effect, which occurs when the pressurized tubes are bent into triangles from their natural linear state. This effect generates internal forces that attempt to return the tubes to their original unbent shape, creating additional stress on the passive unit since the tubes are fixed onto it. 

\begin{figure}[tbp]
  \centering
  \includegraphics[width=\linewidth]{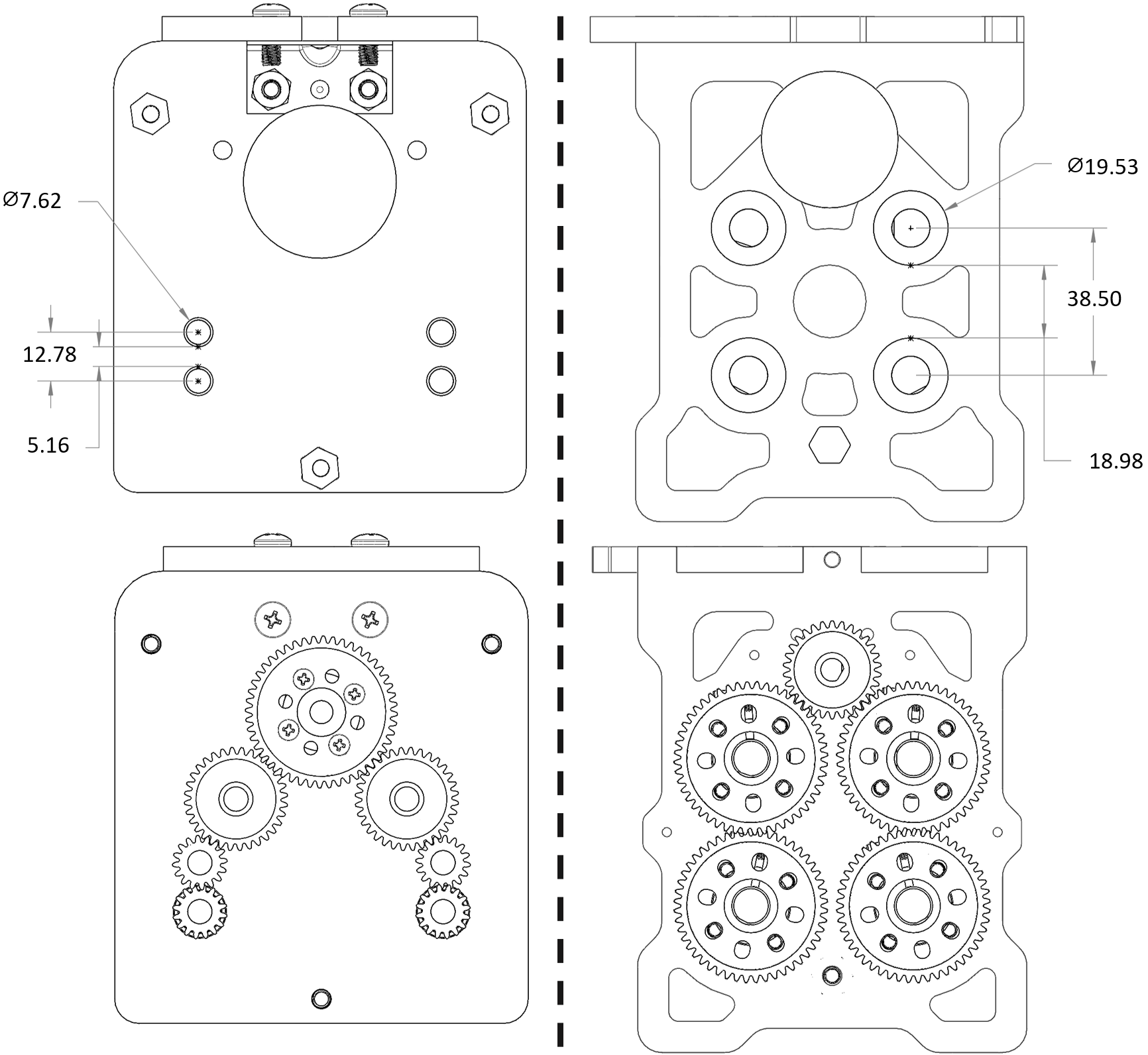}
  \caption{Comparison between the active roller design in \cite{14_usevitch_untethered} (left) and this work (right). The top figures show dimensioning between roller shafts, while the bottom figures show differences between the gear train.}
  \label{fig:active_roller_comparison}
\end{figure}

\begin{figure*}[htbp]
  \centering
  \includegraphics[width=\textwidth]{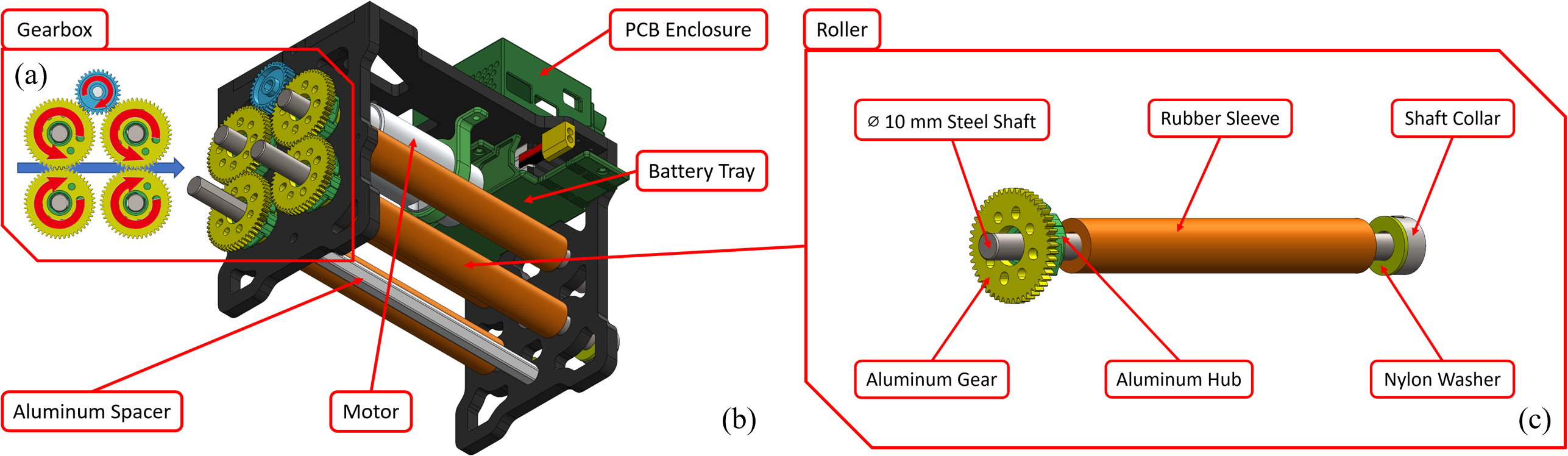}
  \caption{Construction and operation of the active unit and its sub-assemblies. a) Gearbox assembly (with gearbox cover removed) and motion, illustrating how each gear is actuated to produce the linear motion of the tube. b) Architecture of the active roller unit, highlighting its various components. c) Detailed view of the roller-shaft construction.}
  \label{fig:active_architecture}
\end{figure*}

The active roller unit, shown in Fig.~\ref{active_roller_unit}, must perform two different functions crucial to changing the shape of each triangle. First, it creates localized pinch points along the tube that reduce the tube’s bending stiffness and act as effective joints between adjacent sides of a triangle. Second, the active unit must drive along a tube the distance specified by the controller without slipping or jamming so that the structure can change its shape according to the control commands. 

The analysis in \cite{14_usevitch_untethered} shows that the ability of the active roller unit to create pinch points along the tube is primarily governed by two parameters: the gap height between the roller shafts and the roller shaft diameter. The model indicates that minimizing the gap height and roller shaft diameter minimizes the moment required to bend the tube and, therefore, enables the roller unit to translate along the tube more efficiently. The design presented in \cite{14_usevitch_untethered} uses a distance between the centers of the roller shafts of 12.7 mm with a roller shaft diameter of 7.62 mm (6.35 mm steel shaft diameter + 0.6 mm thick rubber sleeve), resulting in a gap of 5.1 mm. Although this configuration effectively minimizes the bending moment, it significantly increases gear train complexity by requiring the inclusion of two idler gears, and results in substantial shaft deflection that limits the maximum operating pressure to 41~kPa.

Because the overall structural strength of the robot increases with tube pressure, a different design tradeoff is adopted in this work. A larger gap height and roller shaft diameter are selected to enable operation at higher pressures, at the expense of increasing the bending moment. As shown in Fig.~\ref {fig:active_roller_comparison}, the current design uses a 38.5~mm center-to-center distance between the two roller shafts with a shaft diameter of 19.5~mm (10 mm steel shaft diameter + 4.75 mm thick rubber sleeve), resulting in a gap height of 19~mm. This design tradeoff allows the truss to operate at higher tube pressures while substantially reducing the complexity of the active roller unit’s gear train.

The construction of the active roller is divided into three subassemblies: the chassis, shown in Fig.~\ref{fig:active_architecture}b, the rollers in Fig.~\ref{fig:active_architecture}c, and the guiding arms.

The chassis consists of three laser-cut polyoxymethylene (POM) components, two sides and a top plate. It provides structural support and mounting interfaces for the PCB, battery, 60 RPM motor (ServoCity 638320), and spherical joint. Each pair of side plates is machined together to ensure concentric roller-shaft holes and precise gear meshing, while the self-lubricating properties of POM further minimize friction and eliminate the need for ball bearings. An aluminum spacer at the base maintains parallel alignment and increases rigidity.

The roller-shafts, shown in Fig.~\ref{fig:active_architecture}c, consist of four precisely machined 10~mm carbon steel shafts (McMaster 8632T38) encased in latex rubber sleeves (McMaster 5234k46) to improve traction with the tube. Aluminum clamping hubs (ServoCity 1301-0016-0010) are used to mount aluminum gears (ServoCity 2302-0014-0048) onto the steel shafts. As illustrated in Fig.~\ref{fig:active_architecture}a, these gears are engaged sequentially to rotate the rollers and generate the linear pulling motion that drives the tube through the unit.

The guiding arms, shown in Fig.~\ref{active_roller_unit}, maintain bisected angles at each truss vertex through a geared mechanism that couples their motion as the tube bends. Their slightly different lengths allow the assembly to fold compactly during stowage.

\begin{figure}[htbp]
  \centering
  \includegraphics[width=\linewidth]{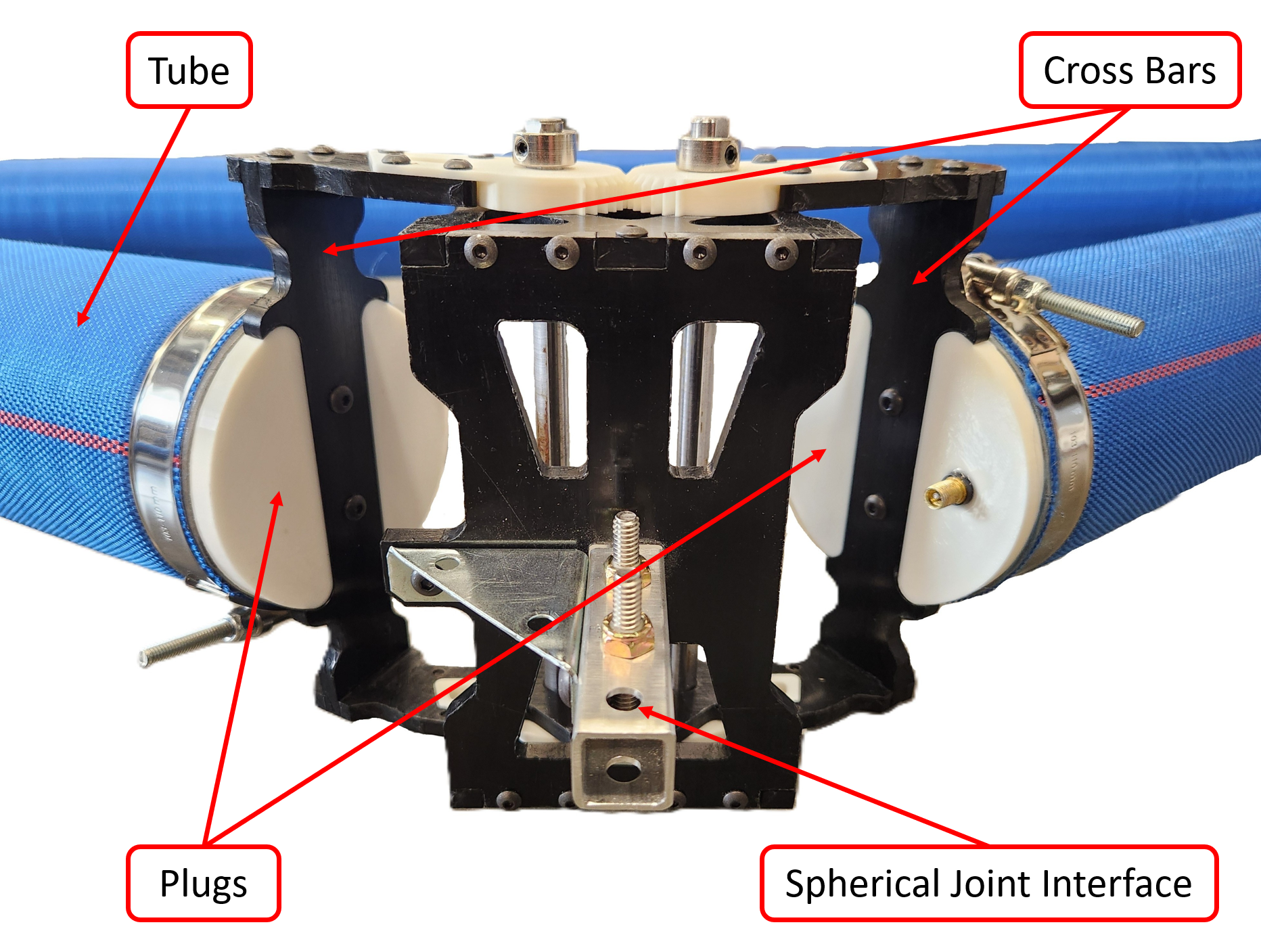}
  \caption{Passive roller unit showcasing the connection between the two ends of the tube along with the plugs, cross bars for fastening the tube to the passive unit, and the interface for connecting to a spherical joint.}
  \label{passive_roller_unit}
\end{figure}

The passive roller unit is a simplified version of the active unit with the primary function of connecting the two ends of the tube, as seen in Fig.~\ref{passive_roller_unit}, so that each triangle is one continuous substructure. This connection is formed by sealing the tube around plugs bolted to crossbars, which link directly to the passive roller arms.

The passive roller unit is composed of the same sub-assemblies as the active unit, only that they are simplified since the passive unit does not require a motor, gearbox, or electronics. Moreover, since the tube does not run through the passive unit, only two shafts without the rubber sleeve are used to mount the guiding arms.

The latest roller unit designs incorporate lighter components and employ topology optimization to achieve substantial weight reductions compared to \cite{14_usevitch_untethered}, as shown in Table~\ref{mass_reduction_active}. These improvements result in a 31\% and 38\% reduction in mass for the active and passive roller units, respectively.

\begin{table}[!t]
\caption{Comparison of Roller Unit and Triangle Masses Between the Current Design and the Design in \cite{14_usevitch_untethered}\label{mass_reduction_active}}
\centering
\footnotesize
\begin{tabular}{|c||c|c|c|}
\hline
\textbf{System} & \textbf{Initial Design \cite{14_usevitch_untethered}} & \textbf{Final Design} & \textbf{Reduction} \\
\hline
Active Unit & 2.83 kg & 1.98 kg & 31\% \\
\hline
Passive Unit & 1.6 kg & 1.0 kg & 38\% \\
\hline
Triangle & 7.26 kg & 4.96 kg & 32\% \\
\hline
\end{tabular}
\end{table}

\subsubsection{Tube Construction}

The primary structural components of the robot are inflatable tubes that must maintain sufficient internal pressure to resist buckling, bending, and twisting under applied loads. When fed through the rollers, the inflated tubes form triangles of varying side lengths and angles, depending on the positions of the active units. 

We selected tube design is an off-the-shelf commercial discharge hose that incorporates an outer Nylon layer for structural integrity and an airtight polyurethane inner lining. We utilize tubes that  measure 3.65 meters in length, 100 millimeters in diameter, and 1.6 millimeters in wall thickness. The tube ends are sealed with custom plugs that are then connected to the cross bars of the passive rollers using two bolts, as depicted in Fig.~\ref{passive_roller_unit}.

The custom plugs were fabricated out of PLA tough using additive manufacturing. Once printed, the plugs were sealed using Dichtol by Diamant to ensure that they are airtight. There are two types of custom plugs: one equipped with a Schrader valve to enable inflation and one with no additional features.

\begin{figure}[tbp]
  \centering
  \includegraphics[width=\linewidth]{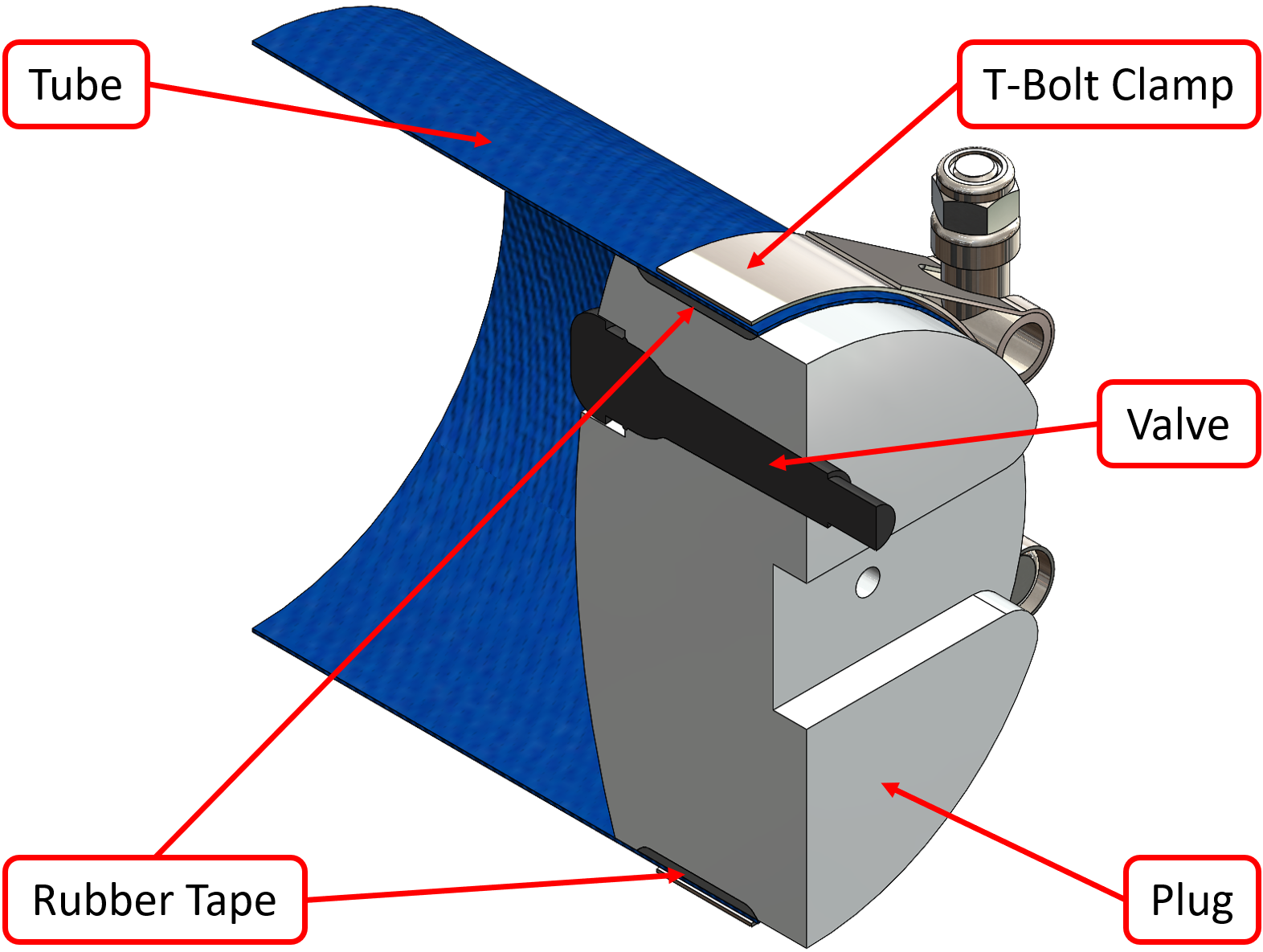}
  \caption{Cross-section along the center of the tube showcasing the seal created by compressing the tube onto the plug using a t-bolt clamp and rubber tape.}
  \label{tube_cs}
\end{figure}

To create an airtight seal between the plug and the tube, a strip of rubber sealing tape is wrapped along the circumference of the plug. A T-bolt clamp is then fastened onto the tube, compressing the tube against the rubber tape. This assembly, shown in Fig. \ref{tube_cs}, creates an air-tight seal that allows the tube to maintain internal pressure.



\subsubsection{Spherical Joint}

The spherical joint serves as the connection point between triangles, enabling free motion at each vertex of the robot and allowing it to achieve a wide range of configurations when changing its shape. For the single octahedron presented previously in \cite{14_usevitch_untethered}, the maximum number of triangles meeting at a corner was 2. In that case, a conventional joint with axial rotation (a combination of pivoting motion with independent rotation of the connected shafts about their longitudinal axes) provides three rotational degrees of freedom: yaw, pitch, and roll. However, in configurations requiring more than two connections at a single vertex, such as the stacked octahedron, a hinge joint with axial rotation becomes inadequate since it can only connect two triangles at a time. To address this limitation, a novel modular spherical joint was developed that is similar in concept and kinematic behavior to the joint in \cite{jeong2018variable}. Although it achieves the same motion, it is different in its mechanical construction, allowing multiple members to connect at a single vertex with a shared virtual center of rotation while preserving the necessary range of motion.

\begin{figure*}[htbp]
  \centering
  \includegraphics[width=\textwidth]{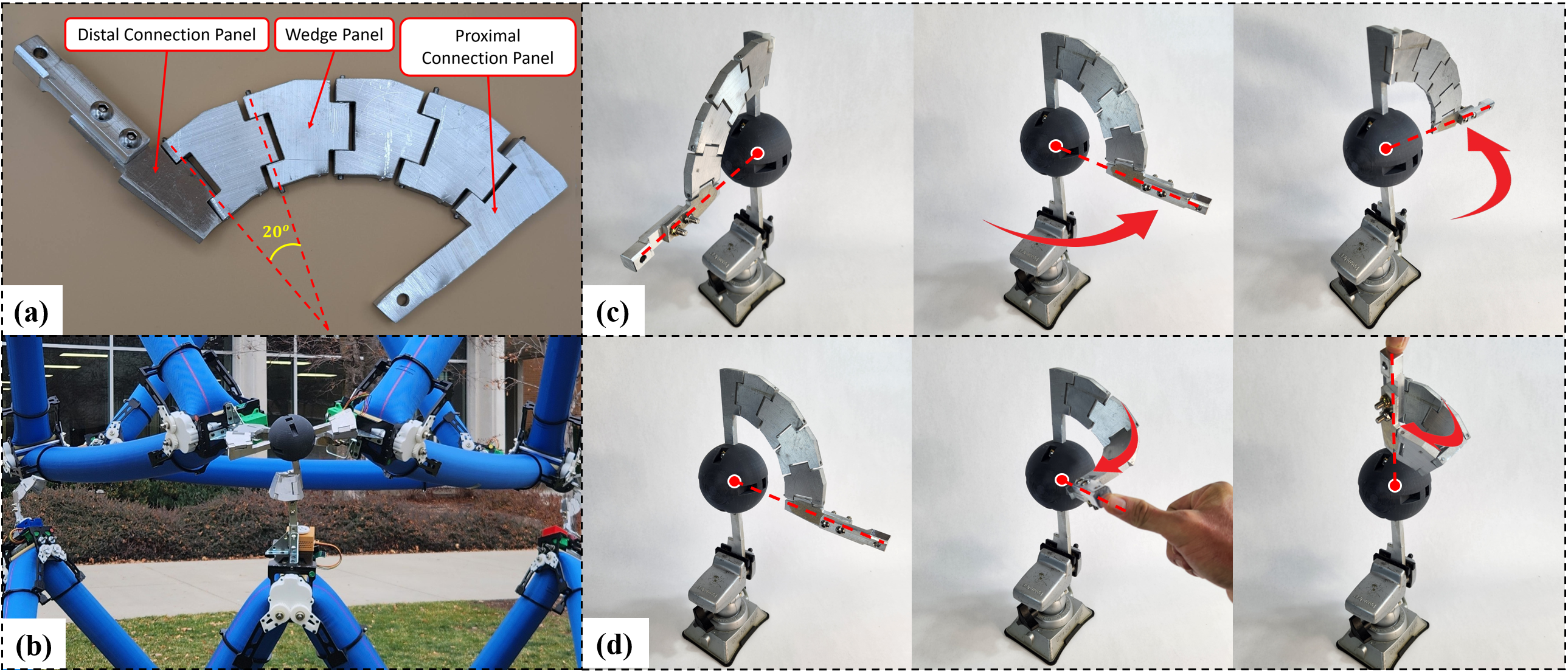}
  \caption{The spherical joint construction, connection, and motion. (a) The arm assembly. (b) A spherical joint connects three triangles. (c) Sweeping motion of the spherical joint with the virtual center highlighted by the red dot with a white border and the center axis of the distal panel by the red dashed line. (d) Curling motion of the spherical joint.}
  \label{fig:spherical_joint}
\end{figure*}

Maintaining a virtual center of rotation is a critical design criterion. Without this alignment, different centers of rotation among the panels would introduce unwanted degrees of freedom that can lead to instability, vibration, and unpredictable motion. Therefore, to ensure consistent motion, the arm design shown in Fig.~\ref{fig:spherical_joint}a consists of four rigid wedge panels connected in series by pinned joints and their edges spaced at $20^\circ$ and oriented toward this virtual center. The distal connection panel connects to the rollers using a clevis pin, shown in \ref{fig:spherical_joint}b, allowing for easy assembly and disassembly. This feature supports the robot’s modularity, enabling an astronaut wearing cumbersome gloves to quickly reconfigure the triangles into various structures as needed. The proximal connection panel secures the joint arm to a central spherical node, which connects all intersecting arms and maintains a virtual center of rotation in the geometric center of the sphere throughout the motion of the panels, as indicated in Fig. \ref{fig:spherical_joint}c and \ref{fig:spherical_joint}d.

While the joint provides sufficient motion for the robotic truss, the effects of altering panel number, angle, and material on joint range of motion and loading behavior should be further explored to understand the potential of this novel joint.

\subsubsection{Electronics and Software}

Each active roller unit is equipped with a custom printed circuit board (PCB). The PCB uses an Arduino Nano, a half-bridge motor driver, a 2.4 GHz radio module, a 3s 1300 mAh Lithium-ion polymer battery, and a 60 RPM planetary DC motor with an attached incremental encoder. Each PCB is mounted onto its respective active roller unit on the top plate of the chassis.

We control the robot through velocity commands for the individual active roller units. These velocities are transmitted via radio over a serial connection to each active unit, where a PID velocity controller executes the commands. Alternatively, the controller can receive target tube lengths to position the units at specified locations.

\subsection{Modeling}

\subsubsection{Robot Definition}
We describe the robot and its kinematics in terms of three variables and their relationships: node positions, edge lengths, and roller positions along the tubes. The robot is represented as a graph $\mathcal{G} = (\mathcal{V}, \mathcal{E})$, where $\mathcal{V}$ denotes the set of nodes and $\mathcal{E}$ the set of edges. The edges are directed such that every three directed edges form a cycle corresponding to a physical tube in the robot (i.e., the first three edges define the first triangle, the next three define the second triangle, and so on).

Node positions are collected in the vector $x \triangleq [{p}_1^\intercal,{p}_2^\intercal,\ldots,{p}_n^\intercal]$, where ${p}_i \in \mathbb{R}^3$ denotes the position of node $i$. The edge-length vector $L$ is defined as:

\begin{equation}
L_k = \left\lVert \mathbf{p}_i - \mathbf{p}_j \right\rVert, 
\quad \forall\, k \in \mathcal{E},\; \{i,j\}_k
\end{equation}

The relationship between the velocity of the nodes and the rate of change of the edge-length vector is given in \cite{14_usevitch_untethered} as:

\begin{equation}
    \dot{L}=R(x) \dot{x}
\end{equation}

where $R(x)$ is the scaled rigidity matrix. 

The corner of each physical triangle in the robot contains a roller unit. We denote the vector of the position of each roller unit along the tube as $d_{all}$. The relationship between the roller unit positions and the edge lengths is given by:

\begin{equation}
    \dot{L}=B_{all}^T\dot{d}_{all}
\end{equation}

Where the matrix $B_{full}$ is the directed incidence matrix of the graph. In practice, one of the three roller units is a passive unit that connects the ends of the tube and therefore cannot move along the tube ($\dot{d}_i=0$). As such, we delete the entries of $d_{all}$ and the columns of $B_{full}$ that correspond to the passive roller units, yielding the following relationship:

\begin{equation}
    \dot{L}=B_{active}^T \dot{d}_{active}
\end{equation}







\subsubsection{Inverse Kinematics}
\label{sec:inverse_kinematics}

We now develop the inverse kinematics used to control the robot. This procedure allows a desired motion to be specified in Cartesian space and computes the active unit velocities required to realize that motion. This formulation builds on \cite{usevitch_triangle_2025}, where a similar approach is used to characterize the workspace of various isoperimetric robots.

In our case, one node is constrained to follow a specified velocity, nodes in contact with the ground are constrained to remain stationary, and the perimeter of each triangle is constrained to remain constant. The remaining node velocities are obtained by solving the following quadratic program with linear constraints over the Cartesian velocity vector of all nodes, $\dot{x}$.

\begin{equation}\label{eq:qp_previous}
\begin{aligned}
\min_{\dot{x}} \quad & \|R(x)\dot{x}\|^2 \\
\text{s.t.} \quad
& A_{\text{move}} \dot{x} = b_{\text{move}}, \quad \text{(move)} \\
& A_{\text{fixed}} \dot{x} = 0, \quad \text{(fixed)} \\
& P R(x) \dot{x} = 0 \quad \text{(perimeter)}
\end{aligned}
\end{equation}

The first ``move'' constraint enforces a prescribed Cartesian velocity on a target node, such that 
$\dot{p}_{\text{move}} = b_{\text{move}}$. 
The matrix $A_{\text{move}} \in \{0,1\}^{3 \times 3n}$ contains one nonzero entry per row, equal to 1, selecting the three Cartesian velocity components of the specified node.  

The second ``fixed'' constraint enforces that a subset of nodes remains stationary. Let $\mathcal F=\{i_1,\ldots,i_m\}\subset\{1,\ldots,n\}$ denote the indices of the fixed nodes. Define the selection matrix $S\in\{0,1\}^{m\times n}$ whose $k$th row is the standard basis vector $e_{i_k}^\top$. This constraint is written as:
\begin{equation}
A_{\mathrm{fixed}}\dot{x} = 0, \qquad 
A_{\mathrm{fixed}} \triangleq S \otimes I_3 .
\end{equation}

The third constraint enforces a constant perimeter for each physical triangle in the robot. Let $P = I_{n_{\text{triangles}}} \otimes [1 \;\; 1 \;\; 1]$, where $n_{\text{triangles}}$ is the number of triangles. Since $R(x)\dot{x}$ gives the vector of edge-length rates, the perimeter constraint is expressed as:
\begin{equation}
P\dot{L}=P R(x)\dot{x} = 0.
\end{equation}

This ensures that the sum of the three edge-length rates for each triangle is zero, preserving the isoperimetric property during motion.

The resulting node velocities are then used to compute the required active unit motions. These actuator velocities are recovered from the Cartesian node velocities using the node-to-roller mapping matrix $B$. Because $B^\top$ is generally non-square, the Moore--Penrose pseudoinverse is used,
\begin{equation}
\dot{d} = B_{\mathrm{active}}^{\dagger\,\top} \dot{x}.
\end{equation}

The pseudoinverse formulation implicitly encodes which rollers are active. The resulting roller displacement or velocity commands are then transmitted to the robot for execution.

\subsubsection{Integration} 
The desired motion is discretized into incremental time steps. At each time step $k$, a constrained optimization problem is solved to obtain the Cartesian velocity $\dot{x}_k$, which is then integrated to update the truss configuration,
\begin{equation}
x_{k+1} = x_k + \Delta t\, \dot{x}_k .
\end{equation}

From this process, either the final configuration $x_{\text{final}}$ or the full sequence of velocities $\{\dot{x}_k\}$ can be extracted.

\section{Results}

This section presents experimental results demonstrating the functional capabilities and motion behaviors of the inflatable truss robot. The results quantify system-level performance, including power consumption, ease of assembly, storage efficiency, and experimentally validate key motions such as squatting and extension, twisting, top-plane manipulation, and locomotion. Together, these results highlight the robot’s modular assembly, storage capabilities, and ability to achieve versatile, task-relevant motions through coordinated geometric reconfiguration.

\subsection{Functional Capabilities}

\subsubsection{Active Roller Unit Strength and Efficiency}

To assess the durability of the roller units, we conducted a test by mounting one of the active units onto a tube pressurized to 221 kPa, 2.9 times the system’s operating pressure of 76 kPa. The tube was maintained at this elevated pressure for two hours, during which the roller unit was kept stationary. During and after the test, the unit was closely monitored for signs of significant elastic deformation and signs of creep; none were observed.

\begin{figure}[t]
  \centering
  \includegraphics[width=\linewidth]{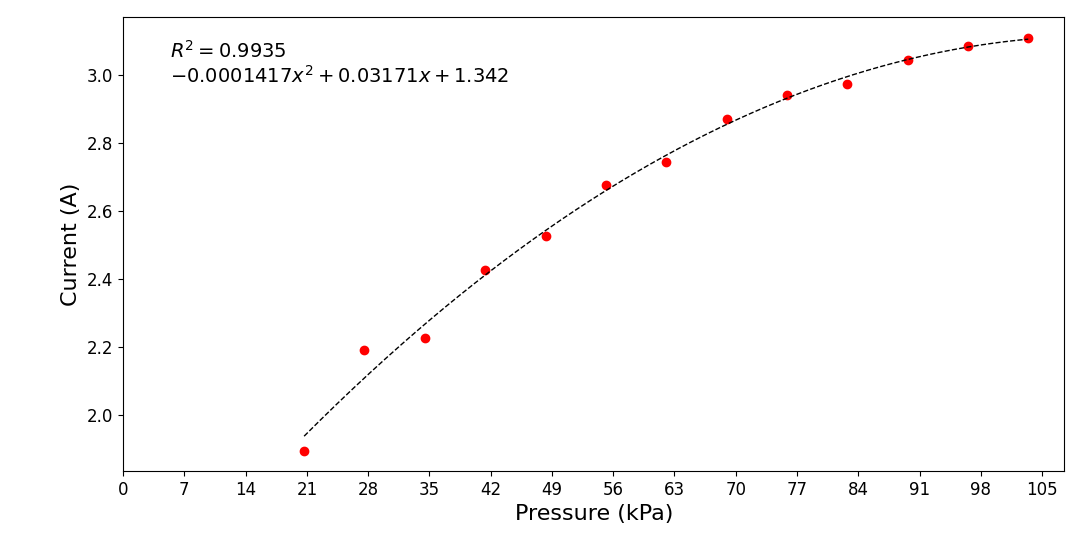}
  \caption{Current to drive the motor along the tube at different pressures. A second-order polynomial fit is shown.}
  \label{current_vs_pressure}
\end{figure}

The amount of electrical current the motor draws to move the rollers is directly related to the pressure in the tubes. As the tube pressure increases, the tube pushes the rollers apart with greater force, increasing friction against the POM chassis. This higher friction requires the motor to exert more torque to actuate the mechanism, leading to greater current draw from the battery. Fig.~\ref{current_vs_pressure} illustrates the motor current draw for an active unit rolling on the base tube of a single octahedron bent at $60^\circ$ and pressurized between 21 and 103 kPa using increments of 7 kPa. At an operating pressure of 76 kPa, the motor draws approximately 2.94 Amps. 

Assuming an average operating current of 11.3 mA for the radio module and 2.94 A for the motor, a 1300 mAh battery provides roughly 26.4 minutes of continuous operation. In practice, the active rollers run only in short bursts (10–15 seconds) to reconfigure the robot, allowing the system to operate significantly longer than the continuous estimate.

\subsubsection{Modularity}

\begin{figure}[htbp]
    \centering
    \subfloat[]{
        \includegraphics[width=0.31\columnwidth]{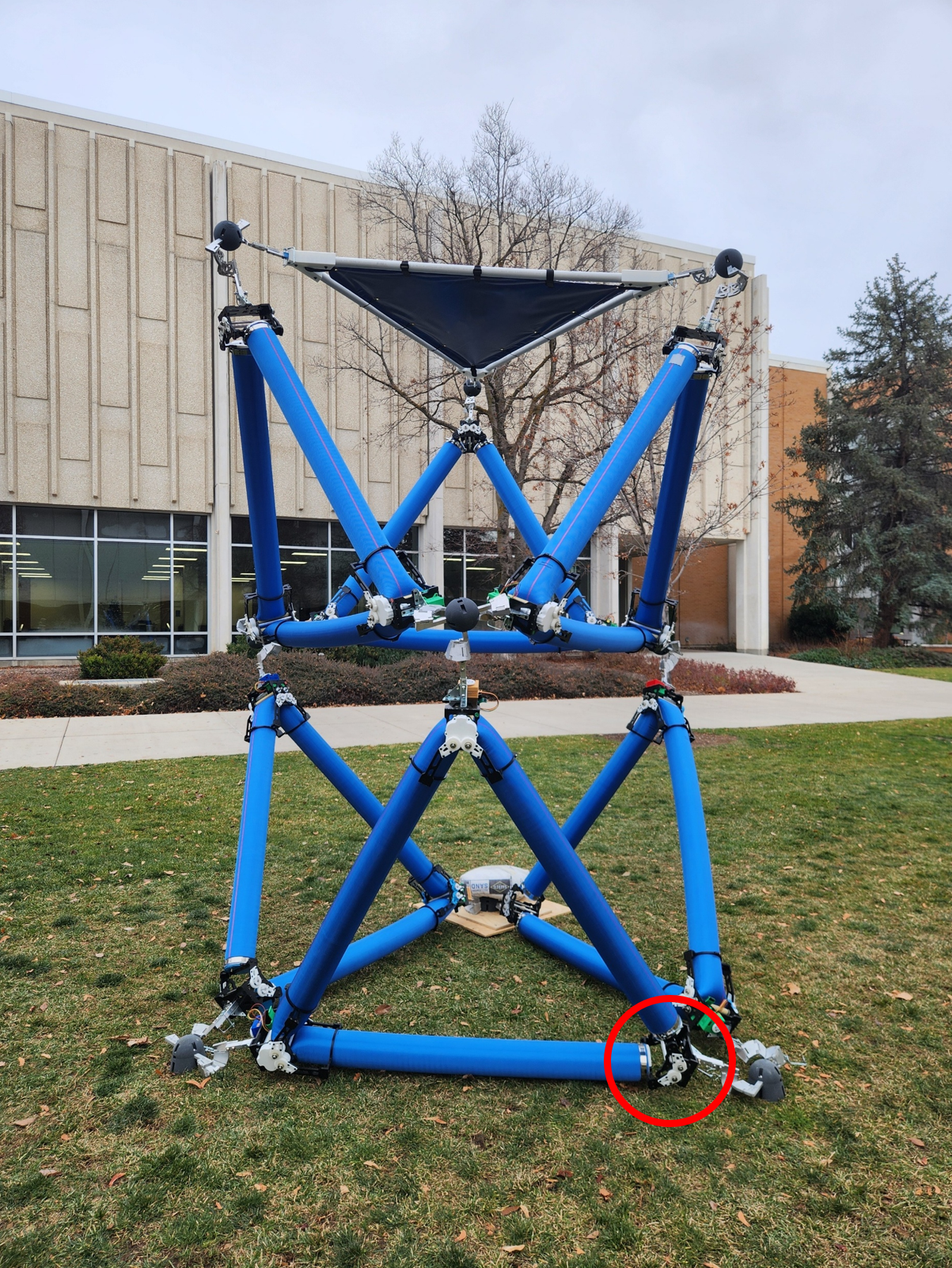}
        \label{fig:modularity_1}
    }
    \subfloat[]{
        \includegraphics[width=0.31\columnwidth]{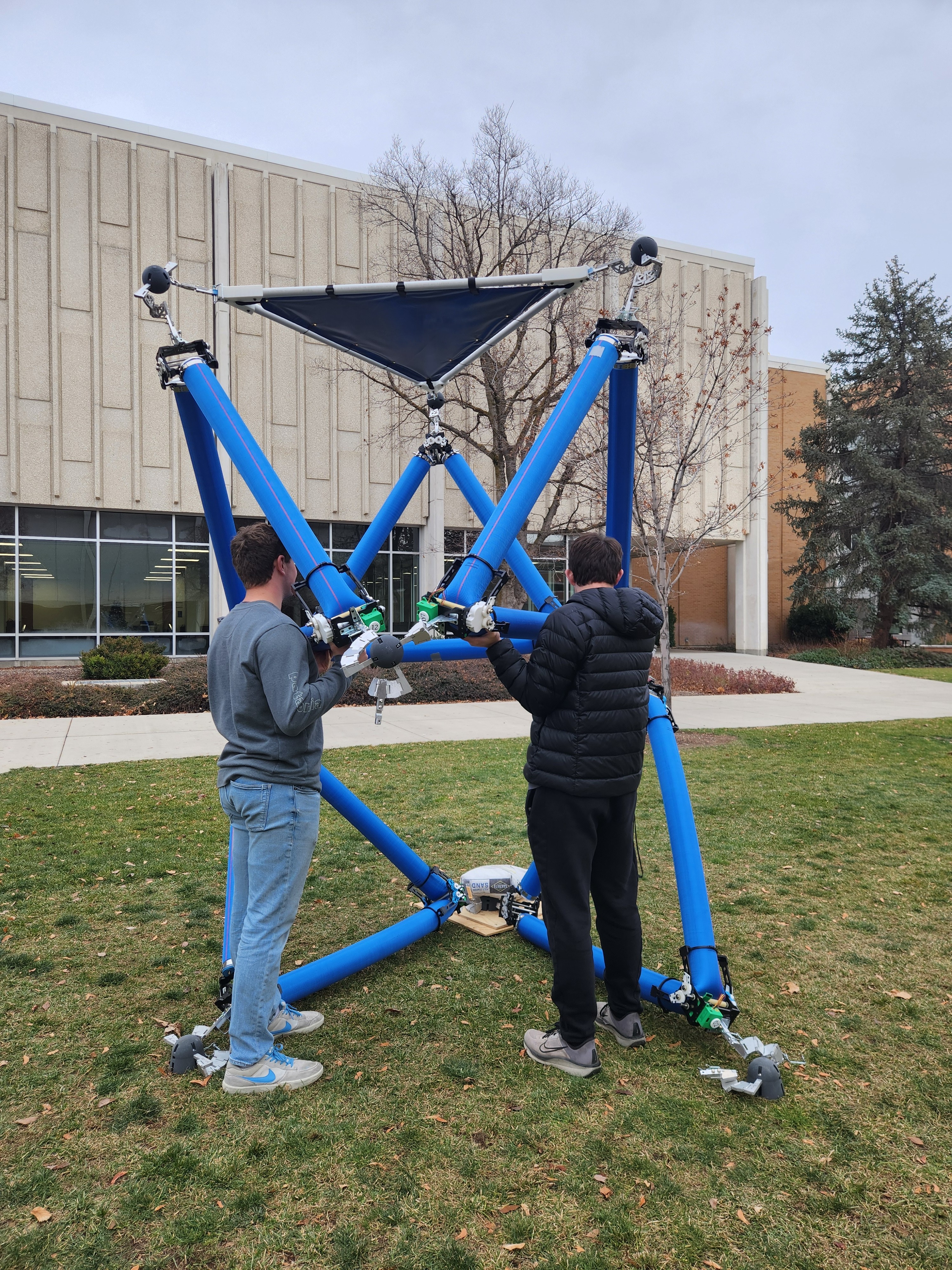}
        \label{fig:modularity_2}
    }
    \subfloat[]{
        \includegraphics[width=0.31\columnwidth]{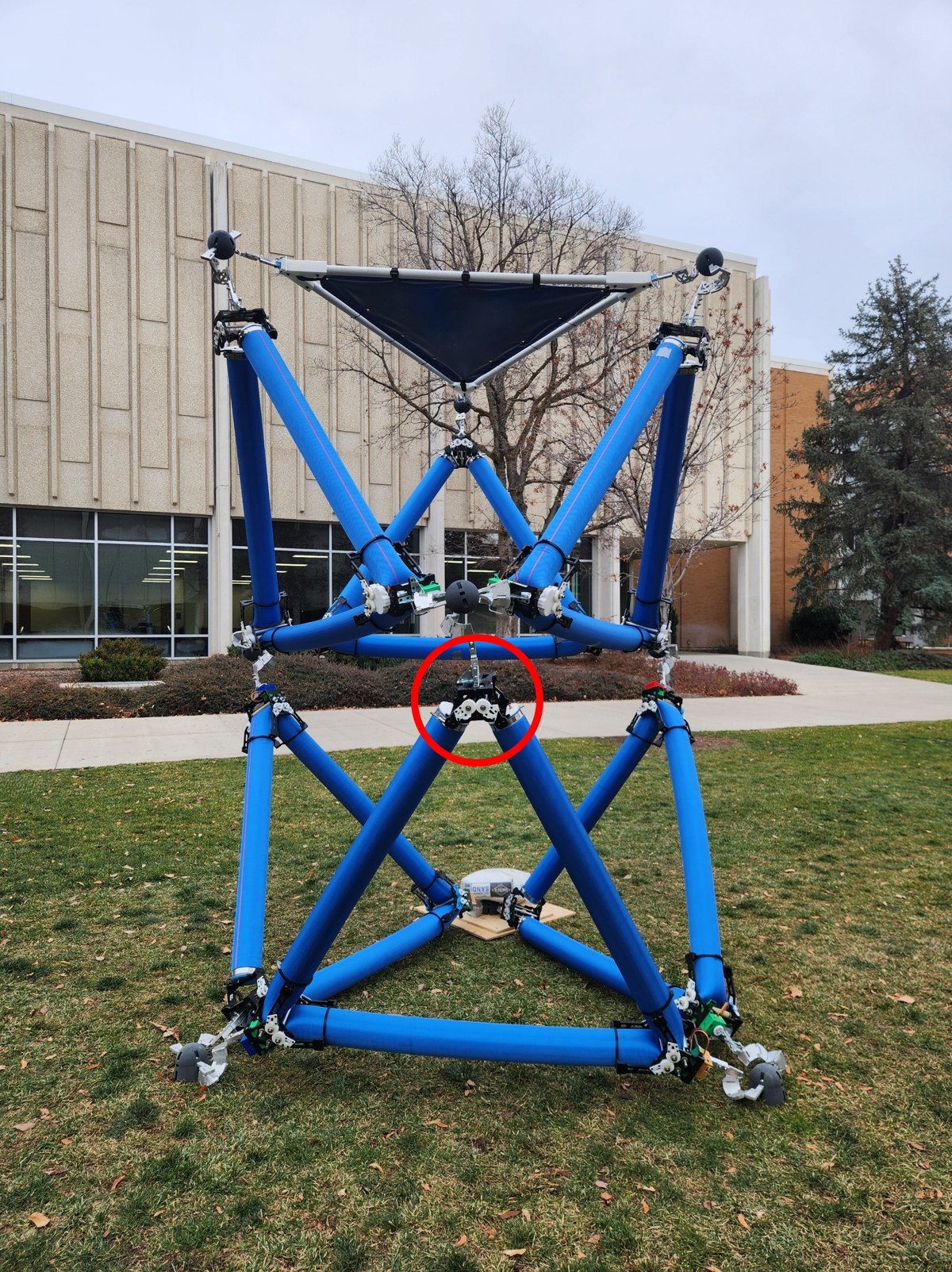}
        \label{fig:modularity_3}
    }
    \caption{Modular reorienting. (a) The passive roller is mounted on the ground. (b) The structure is held while the triangle is rotated. (c) The passive roller is mounted at the top.}
        \label{fig:modularity}
\end{figure}

All spherical joint arms and triangular units are equipped with standardized clevis-pin connections, allowing any triangle to be attached at any location within the structure. Beyond enabling rapid assembly and disassembly, this modularity enables efficient in-field repair. For example, a malfunctioning roller module can be accommodated by reorienting the corresponding triangle, as shown in Fig.~\ref{fig:modularity}. The same procedure may be followed to completely replace a damaged triangle with a functional one. This capability enhances system reliability and supports rapid reconfiguration for storage, deployment, or recovery from technical faults.

\subsubsection{Storage and Deployment Ratio}

A major advantage of inflatable robotic systems is their ability to be stowed in a compact configuration and subsequently deployed to a significantly larger operational volume. This property is especially beneficial for lunar payload transport, where available launch volume is severely constrained. Consequently, the design prioritizes compact stowability to enhance transport efficiency and maximize the stored-to-deployed volume ratio.

\begin{figure}[htbp]
    \centering
    \subfloat[]{
        \includegraphics[width=0.472\columnwidth]{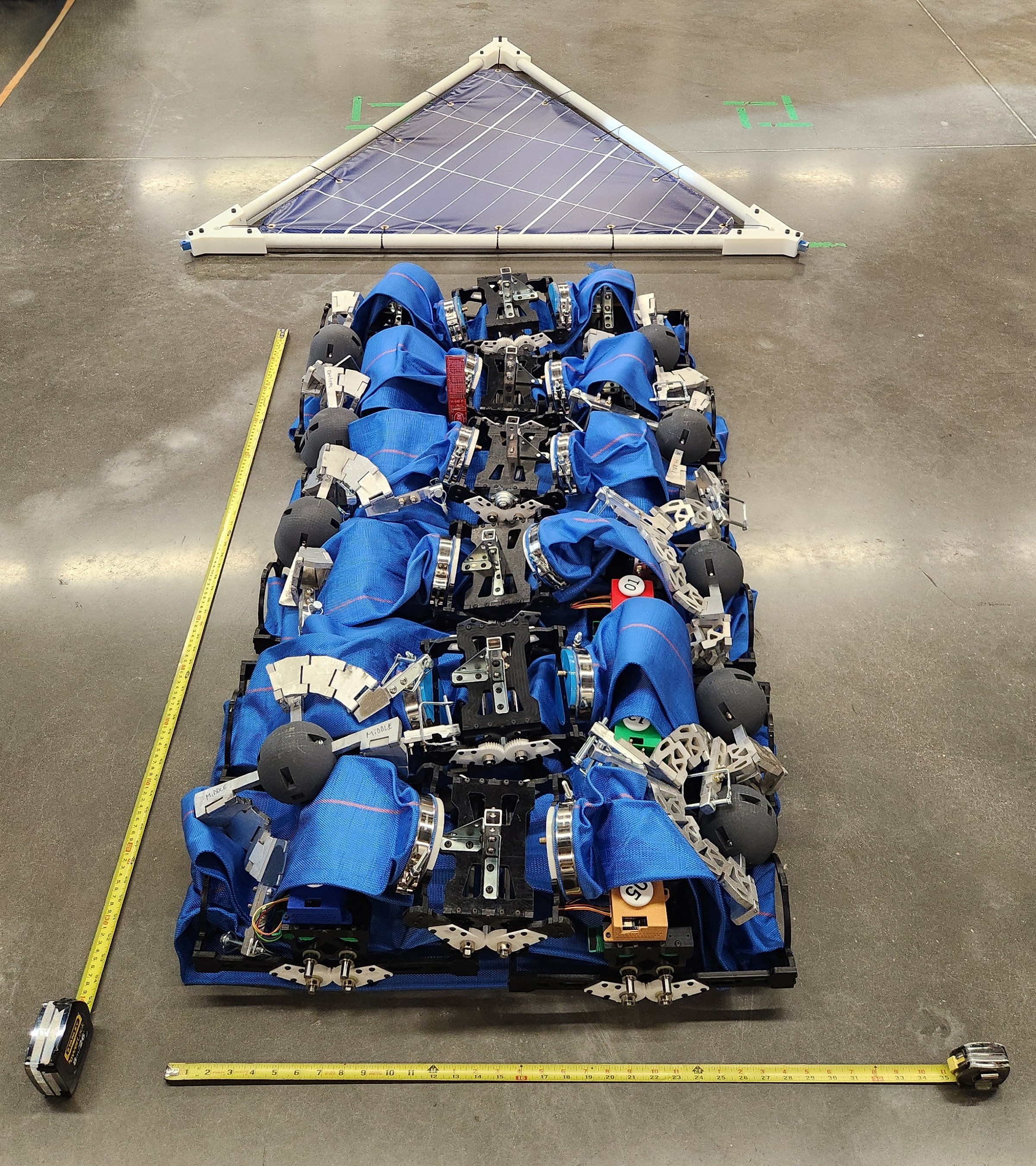}
        \label{fig:6_triangles}
    }
    \subfloat[]{
        \includegraphics[width=0.48\columnwidth]{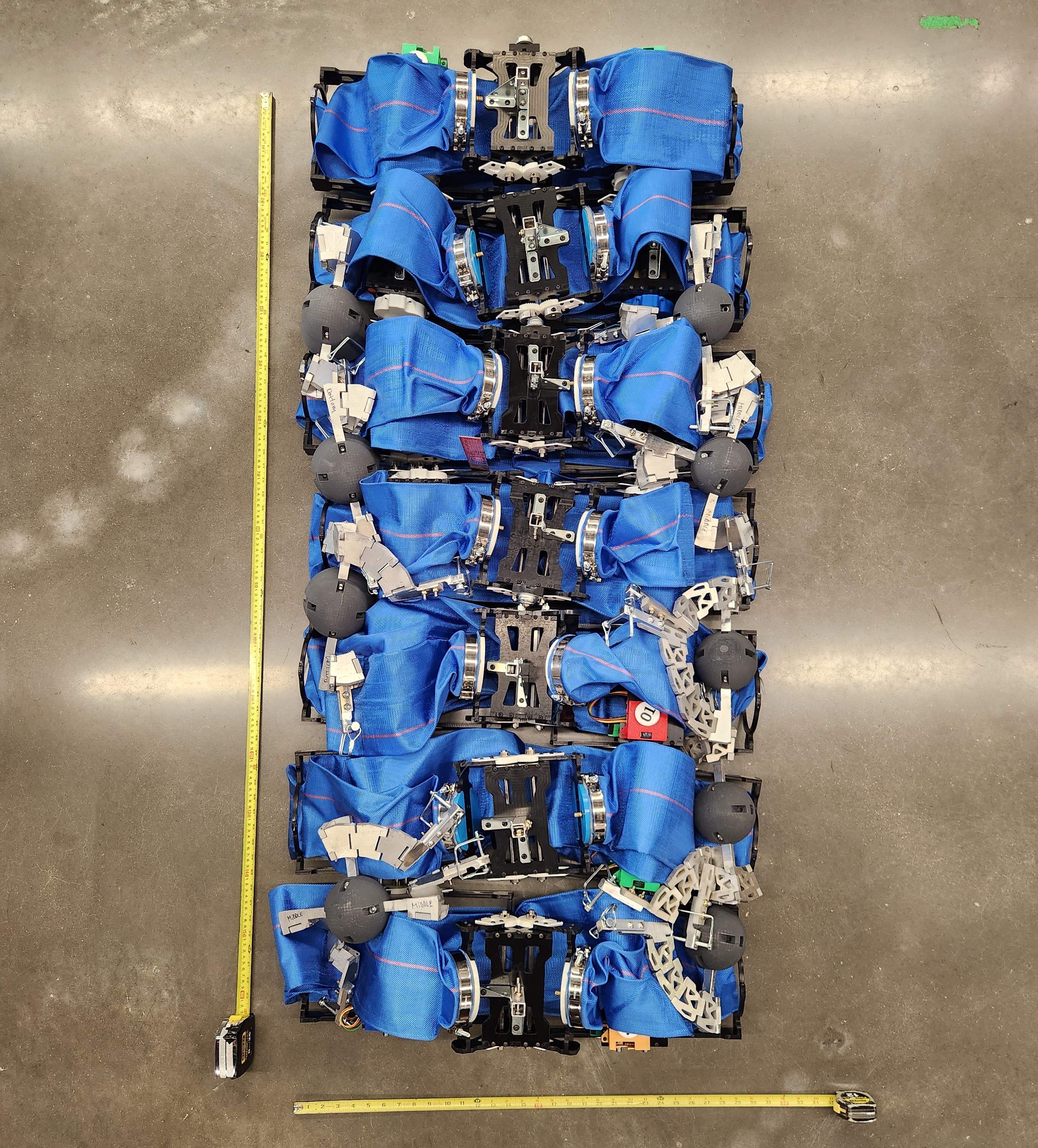}
        \label{fig:7_triangles}
    }
    \caption{Stored configurations. (a) Stored solar array configuration (6 triangles plus solar panel). (b) Stored locomotion configuration (7 triangles).}
    \label{fig:storage}
\end{figure}

Each triangle has a measured stowed volume of 0.043 m\textsuperscript{3} with a footprint of 0.15 m\textsuperscript{2}, while the disassembled (triangles are not connected together) solar panel occupies a stowed volume of 0.038 m\textsuperscript{3} and a footprint of 0.87 m\textsuperscript{2}. As shown in Fig.~\ref{fig:6_triangles}, the solar array configuration can be stowed within a total volume of 0.30 m\textsuperscript{3}, while maintaining a footprint comparable to that of the solar panel, as the triangles may be stacked atop the panel. For the configuration that enables locomotion, the robot can be stowed in a volume of 0.301 m\textsuperscript{3} with a footprint of 1.05 m\textsuperscript{2}, as illustrated in Fig.~\ref{fig:7_triangles}.

The deployed volume of both the solar array and locomotion configurations is estimated by modeling the structure as two regular octahedra with side length 1.8 m, computed as the triangle side length plus the two times the distance from the roller unit to the center of the spherical joint. This yields a deployed volume of 5.5 m\textsuperscript{3}, resulting in a stowed-to-deployed volume ratio of approximately 1:18.3. Currently, deploying the two configurations requires three individuals, motivating future work toward achieving autonomous deployment to reduce astronaut workload and improve operational efficiency in lunar missions.

\subsection{Solar Array Configuration}
\label{sec:motion}

We first present results for a solar array configuration, consisting of two stacked octahedron units where the top triangle is replaced with a rigid panel that represents a solar array. This top surface could represent a solar array, a scientific instrument, or communication equipment that needs to be targeted or moved between different locations and orientations.  This configuration of the robot contains 12 active roller modules as part of 6 inflated triangles. We characterize the possible motions of this configuration. 

The triangles in the bottom octahedron are pressurized to 76 kPa to increase structural strength, while those in the bottom octahedron are pressurized to 41 kPa to reduce torque demands on the motor. For these experiments, we constrain the bottom three nodes with sandbags so that they are fixed relative to the ground. 


\subsubsection{Squatting and Extension}

\begin{figure*}[t]
  \centering
  \includegraphics[width=\textwidth]{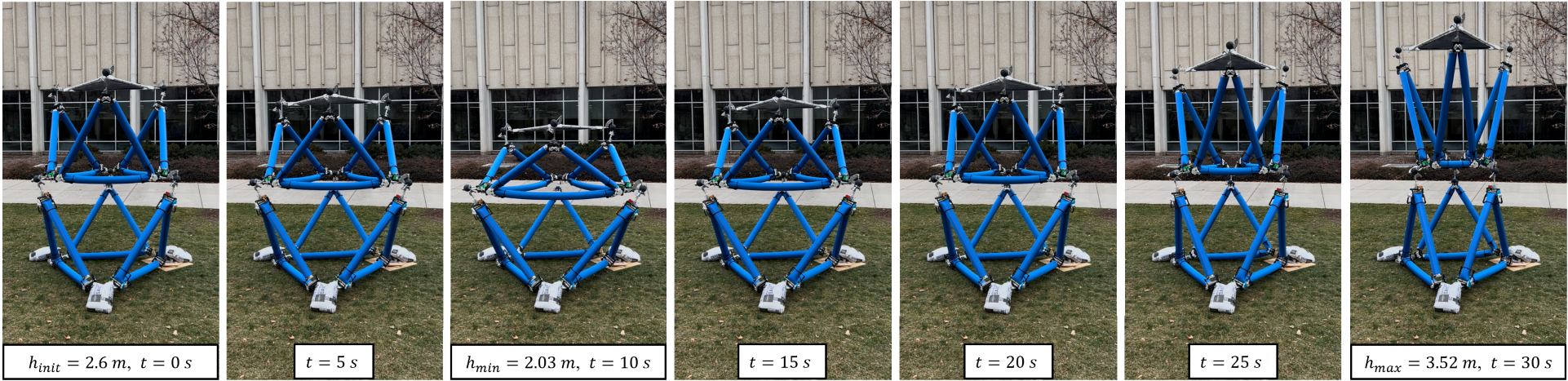}
  \caption{Squatting and extension motion of the ground-fixed truss robot with time and height stamps every 5 seconds. The robot is fixed to the ground due to tilting concerns.}
  \label{fig:squatting_extension}
\end{figure*}

\begin{figure*}[ht]
  \centering
  \includegraphics[width=\textwidth]{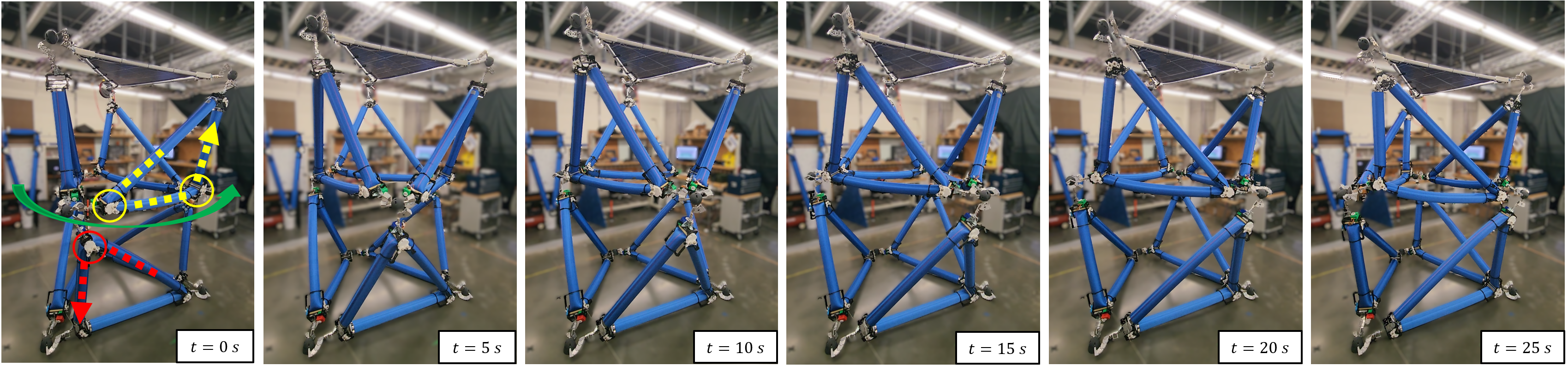}
  \caption{Twisting motion of the truss robot with time stamps at 5 s intervals, showing a $120^\circ$ counterclockwise rotation of each node in the middle plane. Arrows indicate the direction the tube is getting pulled by the circled active units.}
  \label{fig:twisting}
\end{figure*}

By varying the length of each base side of the upper three triangles, the robot can squat down or extend upward, as shown in Fig.~\ref{fig:squatting_extension}. During extension, the triangles transition from equilateral to acute isosceles, whereas during squatting, they transition from equilateral to obtuse isosceles. In the ground-fixed solar array configuration, which constrains the joints in contact with the ground from translating in any direction, the robot achieves a minimum squat height of 2.03 m and a maximum extension height of 3.52 m, corresponding to a vertical range of motion of approximately 1.5 m. Extension is limited by the base side of each triangle, which cannot be shortened further once the two roller units that define it come into contact. Squatting is limited by the maximum allowable lengthening of the triangle base before the triangle inequality constraint is violated. If the lower joints are instead allowed to slide across the ground the bottom octahedron could also contribute to the vertical range of motion. Simulation indicates that the robot can reach a minimum height of 1.22 m and a maximum height of 4.20 m, resulting in a total vertical range of motion of approximately 3.0 m.

\subsubsection{Twisting}

The truss robot can twist both the top and bottom sections together or independently of each other. In Fig.~\ref{fig:twisting}, the tubes in the top triangles are moved such that the right side of each triangle is lengthened and the left side is shortened while maintaining a constant base. The bottom triangles follow the same pattern in the opposite direction, although only one active unit is engaged. This allows either octahedron to twist independently in either the clockwise or counterclockwise direction. Due to these opposing motions, the top plane of the truss does not rotate throughout the motion, while the middle plane rotates by $120^\circ$ counterclockwise. This actuation pattern can be modified such that both octahedra twist in the same direction, or one remains fixed while the other twists, enabling controlled rotation of either the top or middle planes of the structure. The achievable twisting range of motion is limited by the allowable change in triangle side lengths before violation of the triangle inequality constraint.

\subsubsection{Tilting and Sweeping of the Top Plane}
\label{sec:tilt_and_sweep}

\begin{figure*}[t]
  \centering
  \includegraphics[width=\textwidth]{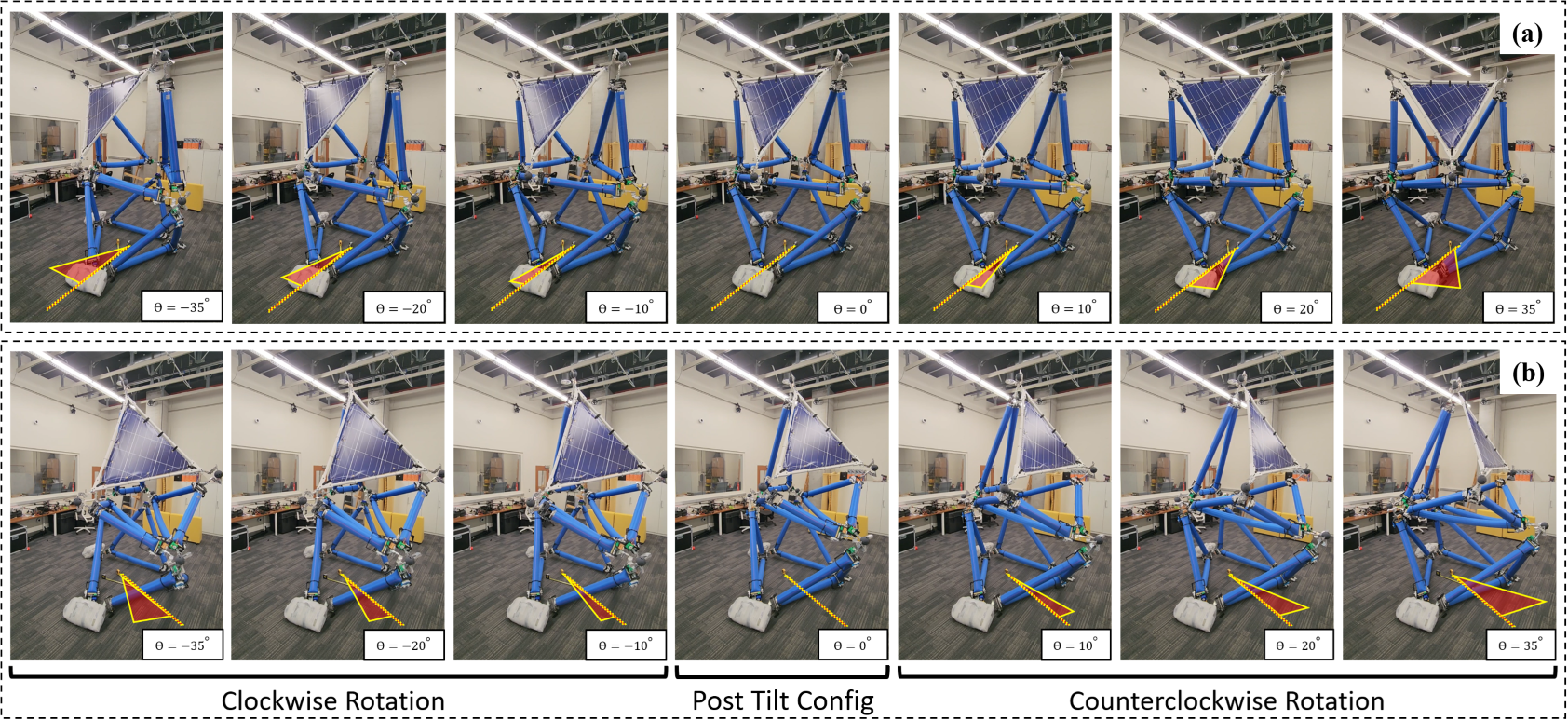}
  \caption{Sweeping motion of the truss robot’s top plane over a) a joint or b) an edge. After tilting, the top plane is swept clockwise from $0^\circ$ to $+35^\circ$ and counterclockwise from  $0^\circ$ to $-35^\circ$ by adjusting triangles in the bottom octahedron.}
  \label{fig:sweeping}
\end{figure*}

\begin{figure}[htbp]
    \centering
    \includegraphics[width=\linewidth]{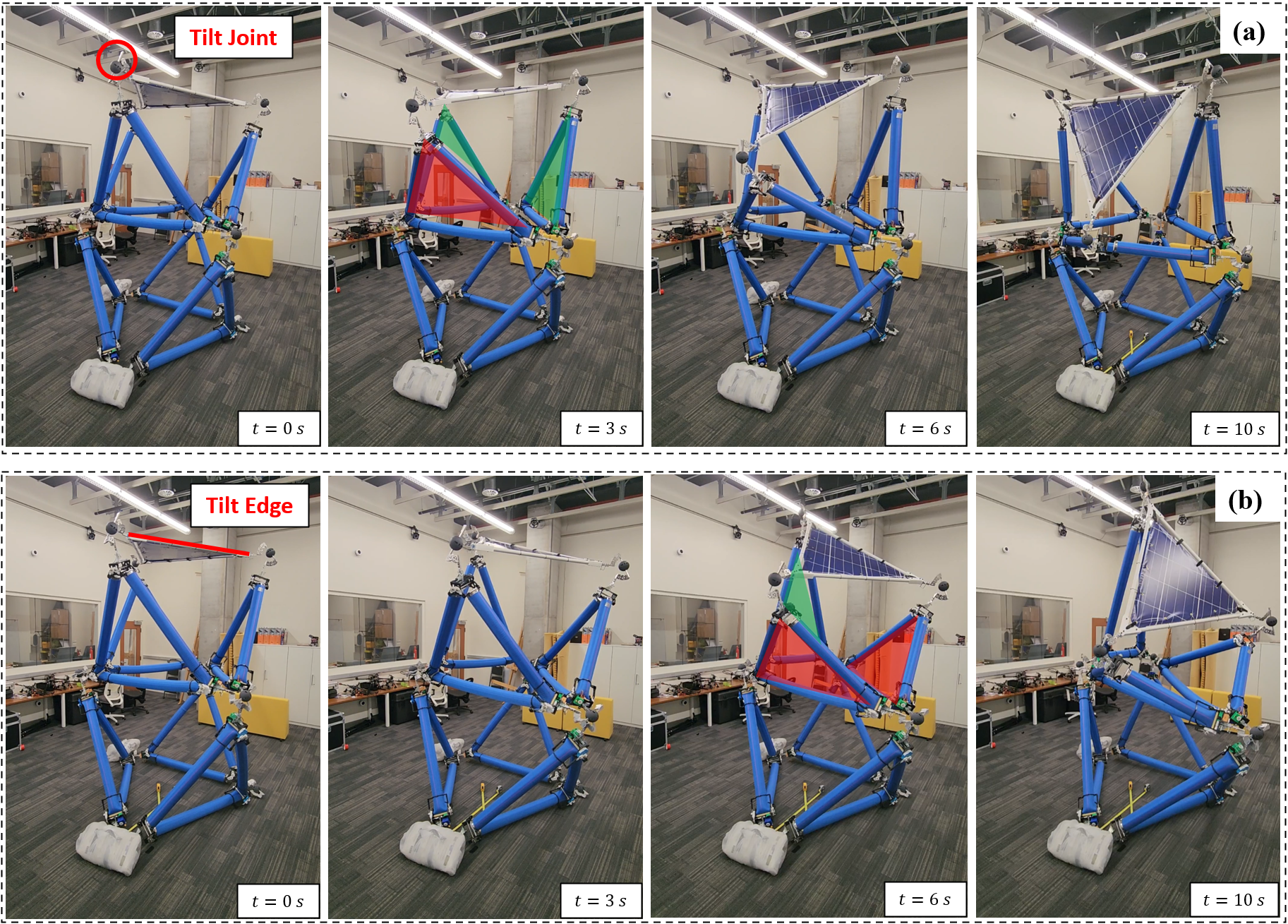}
    \caption{a) Tilting of the truss robot’s top plane about a joint. b) Tilting of the truss robot’s top plane about an edge. Tilting is achieved through selective squatting (red) and extension (green) of the top triangles.}
    \label{fig:tilt}
\end{figure}

The truss robot can tilt and sweep its top plane to adjust both its position and orientation. Tilting may be performed about either a joint or an edge. Tilting about a joint, shown in Fig.~\ref{fig:tilt}a, is achieved by extending the two triangles opposite the tilt joint while squatting the triangle connected to it, resulting in a maximum tilt angle of $65^\circ$ relative to the ground. Tilting about an edge, shown in Fig.~\ref{fig:tilt}b, is achieved by extending the triangle opposite the edge while squatting the two triangles connected to the tilting edge, reaching a maximum tilt angle of $62^\circ$. During both tilting modes, the bottom triangles are adjusted to shift the center of mass toward the center of the structure, reducing the load on the triangles directly beneath the tilt joint or edge.

The sweeping motion is performed by twisting the bottom three triangles using the pattern presented in the previous section. Sweeping achieves a rotation of up to $35^\circ$ in either the clockwise or counterclockwise direction, measured as the angle difference between the projections of the normal to the solar panel onto the ground. The maximum sweeping angle is limited to $35^\circ$ due to the failure mechanisms discussed in \ref{sec:failure}. Although sweeping is therefore limited to $\pm35^\circ$, the robot can tilt its top plane onto any of the three upper joints or edges, which are spaced at $60^\circ$ intervals since the top plane is fixed to be equilateral. Since the robot is able to tilt and sweep about a joint, shown in Fig. \ref{fig:sweeping}a, and about an edge, shown in Fig. \ref{fig:sweeping}b, achieving a sweep angle of $70^\circ$ for each tilt and sweep, the robot achieves an effective rotational coverage of $360^\circ$.
  
\subsection{Locomotion}
\label{sec:locomotion}

The locomotion configuration consists of seven triangles forming two symmetric octahedra connected by a shared central triangle. This symmetry makes this structure more suitable for walking-like motion. 

\begin{figure*}[t]
  \centering
  \includegraphics[width=\textwidth]{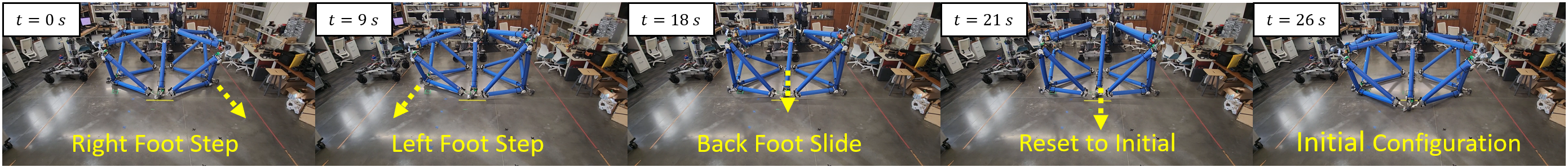}
  \caption{Forward locomotion sequence of the truss robot over two locomotion cycles. The initial reference position is indicated by a yellow line at the foremost spherical joint. The gait consists of advancing the right foot, followed by the left foot, then stepping the rear foot forward, and finally resetting to the initial configuration.}
  \label{fig:locomotion_forward}
\end{figure*}

Forward locomotion is achieved through the command sequence illustrated in Fig.~\ref{fig:locomotion_forward}: step right leg forward, step left leg forward, slide rear center node forward, and reset. These motions are computed using the inverse kinematics detailed in \ref{sec:inverse_kinematics}. To advance a side foot, the top center node is shifted laterally toward one side, transferring the center of mass over the supporting nodes. 

\begin{figure}[htbp]
    \centering
    \includegraphics[width=\linewidth]{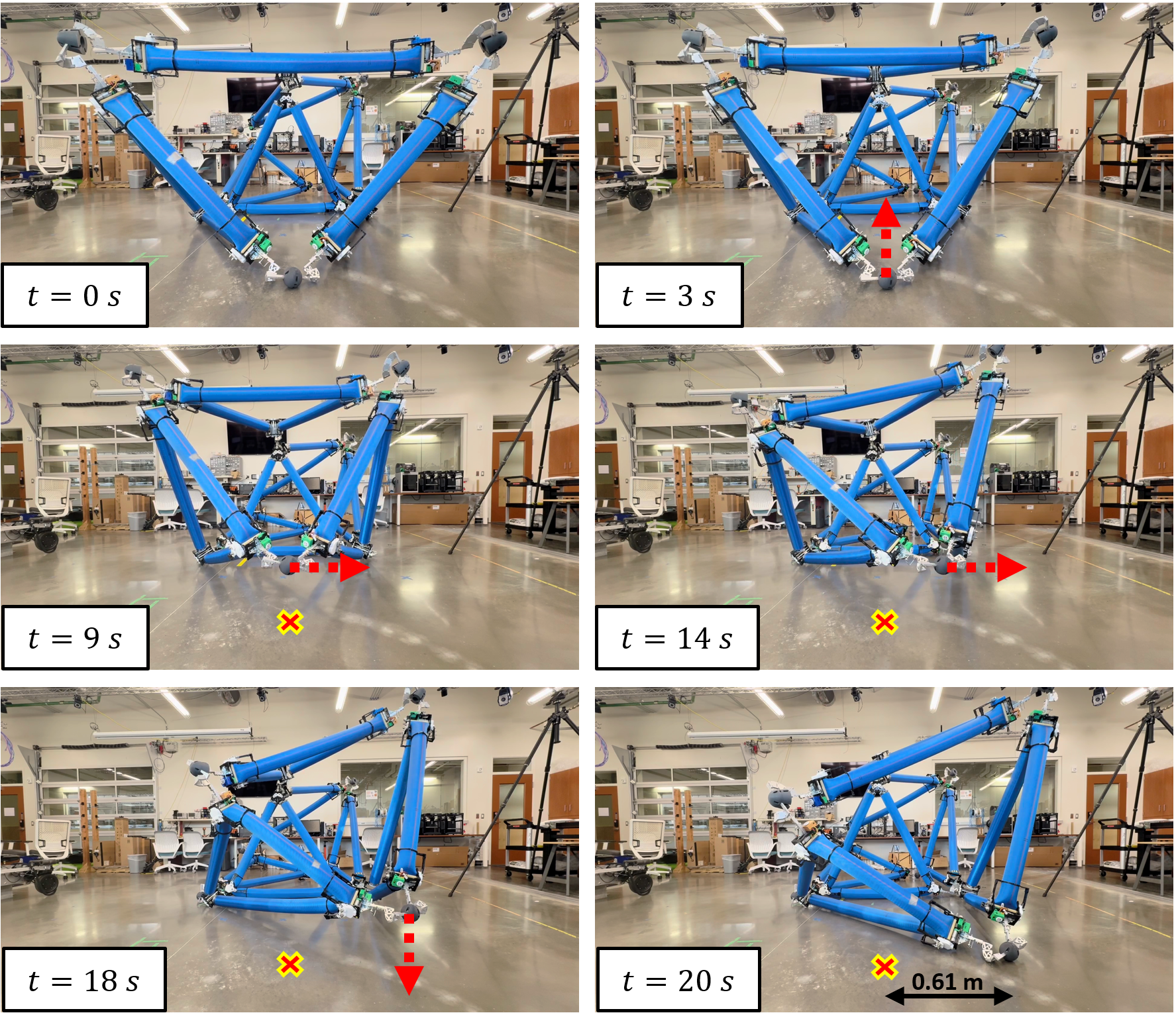}
    \caption{The robot initiates a step by shifting its center of mass to the side opposite the stepping foot over 0–3 s. It then lifts the target node off the ground, translates it forward, and sets it back down. The node’s initial position is marked by the red ×.}
    \label{fig:step}
\end{figure}

Shifting the center of mass allows the node on the opposite side to be lifted, translated forward, and placed back on the ground, as shown in Fig.~\ref{fig:step}, effectively advancing the foot a distance of 0.61 m. The top center node is then shifted to the opposite side, and the same procedure is repeated for the other side foot. Next, the rear center node on the underside of the robot slides forward. The robot then resets to its initial configuration, advancing the foremost node forward before the cycle repeats. Each locomotion cycle produces a net forward displacement of 0.46 m over a duration of 26 s.

\section{Discussion}

\subsection{Failure Mode}

\label{sec:failure}

\begin{figure}[htbp]
  \centering
  \includegraphics[width=\linewidth]{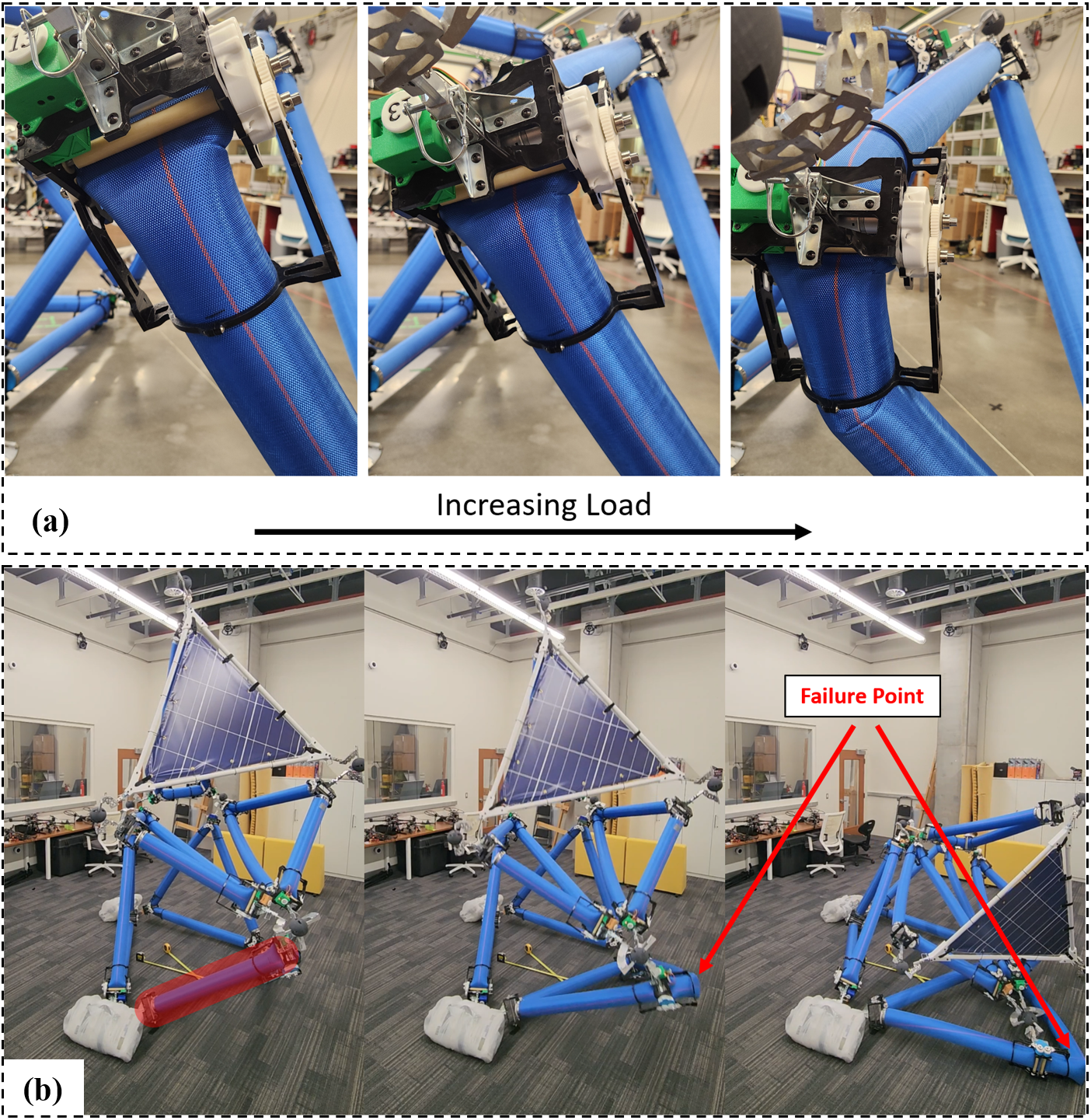}
  \caption{Triangle failure. a) Tube buckling under transverse loading caused by the active roller's guiding arm. b) Collapse of the truss robot with the overloaded triangle and failure point highlighted in red.}
  \label{fig:failure_mode}
\end{figure}

The sweeping motion described in Section~\ref{sec:tilt_and_sweep} is limited to $\pm35^\circ$ as larger rotations generate high bending moments at the interface between the spherical joint arms and the tube. As illustrated in Fig.~\ref{fig:failure_mode}a, these moments cause the arms of the active roller units to press into the tube, introducing stress concentrations. The stress concentrations compromise the structural integrity of the tubes, leading to buckling of the overloaded triangle and eventual collapse of the truss.

Although failure of a single triangle leads to the collapse of the entire robot, shown in Fig. \ref{fig:failure_mode}b, the soft nature of the pressurized tubes attenuates the impact, helping protect the roller units from damage upon ground contact. In addition, the tubes exhibit an inherent springback effect, so only minimal effort is required to re-erect the structure, which can be accomplished by two individuals. In a lunar mission context, these advantages are further amplified by the reduced gravity on the Moon, enabling astronauts to reposition the truss robot with relative ease after collapse.

\section{Conclusion}

This work demonstrates inflatable isoperimetric truss configurations that extend beyond the single-octahedron system presented in \cite{14_usevitch_untethered} through a custom spherical joint enabling three triangles to connect at a single vertex. The resulting structures preserve compact volume, enabling a stowed-to-deployed volume ratio of 1:18.3. Standardized interfaces between triangles, roller units, and joints enable modular construction, reconfiguration, and replacement of damaged components, improving robustness and adaptability. Experimental demonstrations of squatting, twisting, tilting and sweeping, and locomotion show that the system can meaningfully control both position and orientation, supporting tasks such as solar array deployment, crane-like behavior, and terrain interaction. These results highlight the inflatable isoperimetric truss as a promising reconfigurable structure for future lunar missions.

\section{Acknowledgment}

The authors are grateful for the engineering help provided by Ivy Running, Ashleigh Cerve, Logan Yang, Jayden Coe, Henry Smith, Brian Bodily, Simon Charles, Bryce Parkinson, Eli Francom, Jared Berger, and Katie Varela.

\bibliographystyle{IEEEtran}
\bibliography{bibliography}

\section{Biography Section}

{
\setlength{\parskip}{0pt}
\setlength{\baselineskip}{0.95\baselineskip}


\vspace{-36pt}
\begin{IEEEbiographynophoto}{Mihai Stanciu}
received the B.S. degree in mechanical engineering from Brigham Young University in 2024, Provo, Utah, where he is currently working toward the M.S. degree in mechanical engineering with the Compliant Mechanisms and Robotics Lab. His research interests include soft robotics design, optimization of mechanical components, and manufacturing. 
\end{IEEEbiographynophoto}
\vspace{-36pt}
\begin{IEEEbiographynophoto}{Isaac Weaver}
received the B.S. degree in mechanical engineering from Brigham Young University in 2024, Provo, Utah, where he is currently working toward the M.S. degree in mechanical engineering with the Compliant Mechanisms and Robotics Lab. His research interests include soft robotics simulation and modeling, compliant mechanism design, and product design. 
\end{IEEEbiographynophoto}
\vspace{-36pt}
\begin{IEEEbiographynophoto}{Adam Rose}
received the B.S. degree in mechanical engineering from Brigham Young University in 2024, Provo, Utah, where he is currently working toward the Ph.D. degree in mechanical engineering with the Compliant Mechanisms and Robotics Lab. His research interests include compliant mechanism design and lower-limb prostheses design. 
\end{IEEEbiographynophoto}
\vspace{-36pt}
\begin{IEEEbiographynophoto}{James Wade}
is currently pursuing the B.S. degree in mechanical engineering from Brigham Young University. His research interests include robotics and controls. 
\end{IEEEbiographynophoto}
\vspace{-36pt}
\begin{IEEEbiographynophoto}{Kaden Paxton}
is currently pursuing the B.S. degree in mechanical engineering from Brigham Young University. His research interests include robotics and mechanical design. 
\end{IEEEbiographynophoto}
\vspace{-36pt}
\begin{IEEEbiographynophoto}{Chris Paul}
received the B.S. degree in mechanical engineering in 2024 from Brigham Young University. His research interests include soft robotics and PCB design.
\end{IEEEbiographynophoto}
\vspace{-36pt}
\begin{IEEEbiographynophoto}{Spencer Stowell} received the B.S. degree in mechanical engineering from Brigham Young University in 2024, Provo, Utah, where he is currently working toward the Ph.D. degree in mechanical engineering with the Compliant Mechanisms and Robotics Lab. His research interests include soft robotics simulation and modeling, machine learning, and mechanical optimization of lower-limb prostheses. 
\end{IEEEbiographynophoto}
\vspace{-36pt}
\begin{IEEEbiographynophoto}{Nathan Usevitch}
received the B.S. degree in Mechanical Engineering from Brigham Young University in 2015, and the M.S. and Ph.D. degrees in mechanical engineering from Stanford University in 2017 and 2020, respectively. He is currently an assistant professor with the Department of Mechanical Engineering at Brigham Young University. His research interests include the design of novel robotic systems, soft robotics, and modular robotics. 
\end{IEEEbiographynophoto}
}
\end{document}